%% file: acl_latex.tex
\documentclass[11pt]{article}

\usepackage[preprint]{acl}

\usepackage{times}
\usepackage{latexsym}

\usepackage[T1]{fontenc}

\usepackage[utf8]{inputenc}

\usepackage{microtype}
\usepackage{multirow}
\usepackage{makecell}
\usepackage{amsmath}
\RequirePackage{graphicx}
\RequirePackage{amsmath}
\RequirePackage{amssymb}
\RequirePackage{booktabs}
\usepackage{listings}
\usepackage{wrapfig}
\usepackage{xcolor}
\usepackage{colortbl}
\usepackage{enumitem}
\usepackage{academicons}
\usepackage{hyperref}

\definecolor{mycmd}{RGB}{0,90,160}
\definecolor{colGreen}{HTML}{70AD47}  
\definecolor{colRed}{HTML}{E06666}    
\definecolor{colGold}{HTML}{F1C232}   
\definecolor{colBlue}{HTML}{6D9EEB}   

\usepackage{booktabs}
\usepackage{tabularx}
\usepackage[table]{xcolor}
\usepackage{pifont}   
\usepackage{graphicx} 

\newcommand{\cmark}{\textcolor{green!55!black}{\ding{51}}} 
\newcommand{\xmark}{\textcolor{red!65!black}{\ding{55}}}   
\definecolor{cvprblue}{rgb}{0.21,0.49,0.74}

\usepackage{inconsolata}

\title{
    \raisebox{-0.35cm}{\includegraphics[width=1.1cm]{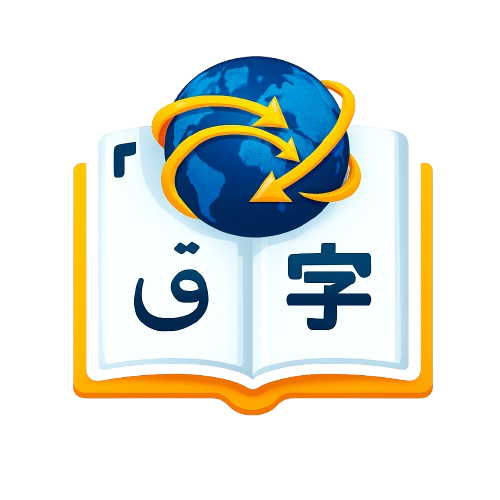}}  
    \hspace{0.01cm} \textbf{\Large DocAtlas: Multilingual Document Understanding \\ Across 80+ Languages}
}

\author{ 
 Ahmed Heakl$^{\spadesuit}$,
 Youssef Mohamed$^{\spadesuit}$,
 Abdullah Sohail$^{\spadesuit}$,
 Rania Elbadry$^{\spadesuit}$ \\
 \textbf{Ahmed Nassar$^{\heartsuit}$, 
 Peter W. J. Staar$^{\heartsuit}$, 
 Fahad Shahbaz Khan}$^{\spadesuit}$ \\ 
 \textbf{Imran Razzak$^{\spadesuit}$, 
 Salman Khan}$^{\spadesuit}$  \\
 $^{\spadesuit}$MBZUAI \quad $^{\heartsuit}$IBM Research \\
 \href{https://huggingface.co/datasets/ahmedheakl/docatlas_instruct}{%
   \raisebox{-0.15em}{\includegraphics[height=0.9em]{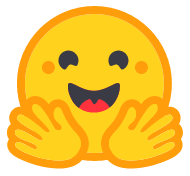}}%
   \hspace{0.3em}\texttt{ahmedheakl/docatlas\_instruct}}
   \hspace{0.8em}
 \href{https://github.com/ahmedheakl/DocAtlas}{%
   \raisebox{-0.15em}{\includegraphics[height=0.9em, trim=50 50 50 50, clip]{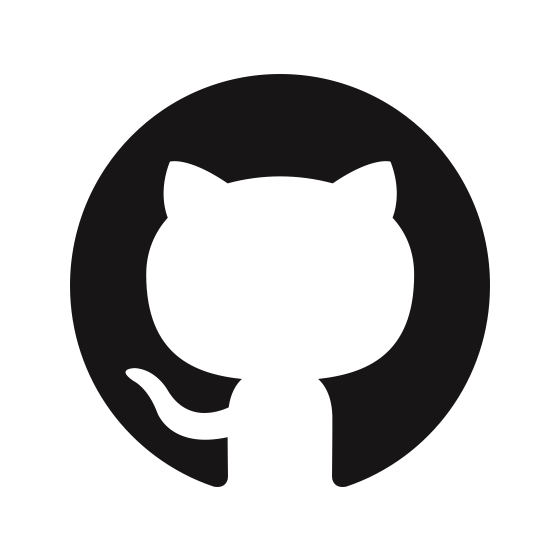}}%
   \hspace{0.3em}\texttt{ahmedheakl/DocAtlas}}
}

\begin{document}
\twocolumn[{
\renewcommand\twocolumn[1][]{#1}
\maketitle
\begin{center}
\captionsetup{type=figure}
\includegraphics[width=\textwidth]{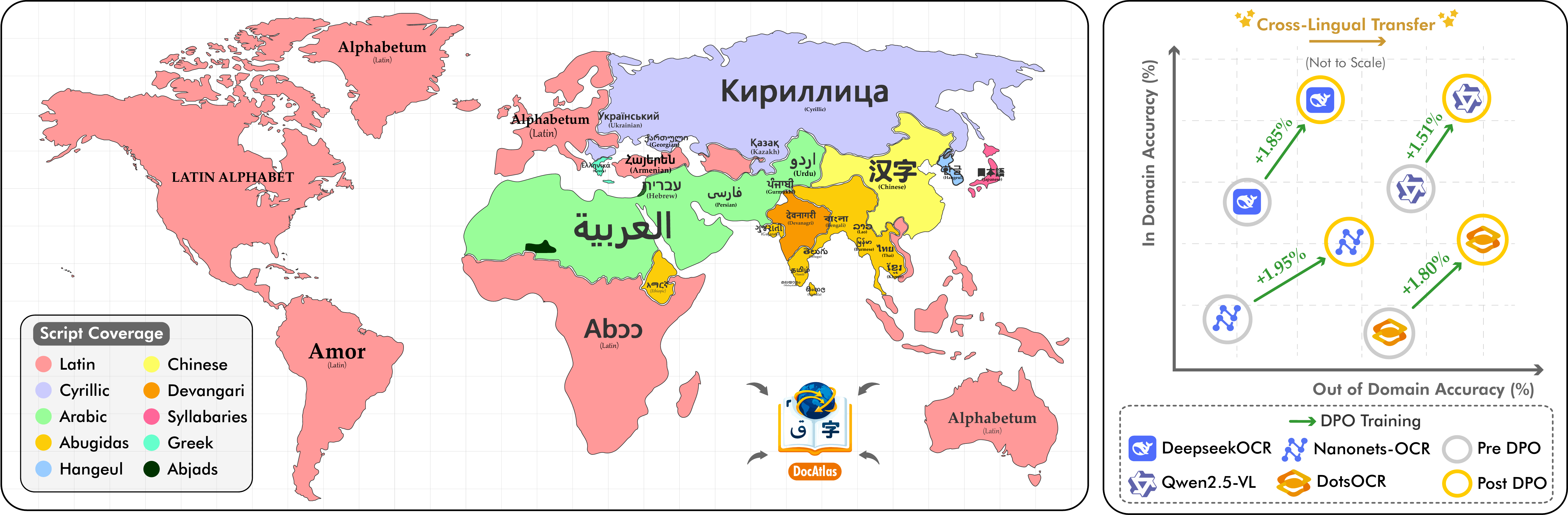}
\vspace{-0.5em}
\captionof{figure}{\textbf{Overview of the DocAtlas framework.}
(\textbf{Left}) Global script coverage across 80+ languages spanning 10 writing systems, illustrating the geographical and typological diversity of the corpus. 
(\textbf{Right}) Cross-lingual transfer performance after DPO training, showing consistent gains in both in-domain and out-of-domain accuracy across major OCR and vision-language models.}
\label{fig:teaser}
\end{center}
}]
\input{sec/0_abstract}
\input{sec/1_intro}
\input{sec/2_related}
\input{sec/3_methods}

\input{sec/4_experiments}

\input{sec/5_results}

\input{sec/6_conclusion}
\input{sec/7_limitations}
\bibliography{custom}
\input{sec/X_suppl}

\end{document}

%% file: sec/0_abstract.tex
\begin{abstract}
Multilingual document understanding remains limited for low-resource languages due to scarce training data and model-based annotation pipelines that perpetuate existing biases. We introduce \textbf{DocAtlas}, a framework that constructs high-fidelity OCR datasets and benchmarks covering 82 languages and 9 evaluation tasks. Our dual pipelines, differential rendering of native DOCX documents and synthetic \LaTeX-based generation for right-to-left scripts produce precise structural annotations in a unified \emph{DocTag} format encoding layout, text, and component types, without learned models for core annotation. Evaluating 14 state-of-the-art models reveals persistent gaps in low-resource scripts. We show that Direct Preference Optimization (DPO) using rendering-derived ground truth as positive signal achieves stable multilingual adaptation, improving both in-domain (+1.9\%) and out-of-domain (+1.8\%) accuracy without measurable base-language degradation, where supervised fine-tuning degrades out-of-domain performance by up to 21\%. Our best variant, DocAtlas-DeepSeek, improves +1.7\% over the strongest baseline.
\end{abstract}

%% file: sec/1_intro.tex
\section{Introduction}
\label{sec:intro}

Despite recent advances in vision-language models (VLMs), multilingual document understanding\footnote{We use \emph{document understanding} to denote the full pipeline from page image to structured output encompassing text, layout, tables, formulas, charts, and reading order, extending beyond character-level OCR.} remains challenging due to the scarcity of high-quality training data across diverse scripts and languages~\cite{liu2024ocrbench,xu2022xfund,xu2021layoutxlm}. While models achieve strong performance on English documents, extending this capability to low- and medium-resource languages is hindered by limited annotated data.

Current dataset construction approaches face critical limitations. Manual annotation~\cite{jaume2019funsd,xu2022xfund} provides high-quality labels but does not scale beyond a handful of languages. Synthetic generation~\cite{yim2021synthtiger,journet2017doccreator} avoids human labor but requires extensive per-script configuration and struggles with complex structures such as nested tables or authentic formatting. Model-based pipelines~\cite{pfitzmann2022doclaynet,paddleocrvl,li2022pp} use pre-trained models to label documents, creating circular dependency where annotation quality is bounded by existing model performance. This perpetuates bias: models trained on English produce annotations that train the next generation of English-centric models. Rendering-based approaches~\cite{weber2023wordscape,dit700} sidestep learned detectors by extracting structure from document source files, but suffer from rendering drift due to lossy format conversion (e.g., LibreOffice), lack geometric alignment between text and bounding boxes, and provide no coverage for right-to-left scripts or structured chart annotation.

We introduce \textbf{DocAtlas}, a pipeline for generating multilingual OCR datasets with model-free structural annotations that extracts ground truth directly from document sources through \emph{differential rendering}. Unlike single-pass colorization, pixel-wise subtraction of colorized and standard renderings disambiguates injected annotations from pre-existing document colors, yielding precise bounding boxes without learned detectors. Results are serialized in a unified DocTag format jointly encoding component type, geometry, and text content, enabling multi-task supervision across all languages. To address the underrepresentation of right-to-left (RTL) scripts, which suffer from PDF parser failures in bidirectional text, we implement a complementary synthetic pipeline that converts structured sources (EPUB, HTML, XML) into PDFs using \LaTeX\ with explicit bidirectional control, generating 52K additional pages with the same annotation precision. Optional metadata enrichment (e.g., figure classification, page attributes) may use auxiliary models, but all core DocTag annotations are fully model-free.

Combining both pipelines yields a 360K-page corpus across 82 languages and a difficulty-stratified benchmark of 5,862 pages spanning 9 evaluation tasks: end-to-end page parsing, text recognition, table extraction, formula transcription, chart parsing, reading order, and three format-specific subtasks (chart$\to$HTML, formula$\to$\LaTeX, table$\to$HTML). We evaluate 14 state-of-the-art models, revealing that low-resource scripts see 40--60\% accuracy drops, structured extraction saturates at 73\% TEDS regardless of language, and chart parsing sharply separates OCR-specialized systems from general VLMs. We further compare adaptation strategies and find that DPO achieves stable cross-lingual transfer (+1.8\% accuracy, $<$3\% base-language degradation) where supervised methods exhibit catastrophic forgetting (up to 21\%), with QKV-only LoRA providing optimal gain-preservation balance. Our contributions are:

\begin{itemize}[leftmargin=*,itemsep=1pt,topsep=2pt]
    \item A differential rendering pipeline producing model-free annotations from 307K documents across 82 languages, addressing five limitations of prior rendering-based approaches (\S\ref{sec:pipeline_a}).
    \item A synthetic RTL pipeline generating 52K pages with precise annotations for underrepresented bidirectional scripts (\S\ref{sec:pipeline_b}).
    \item A difficulty-stratified multilingual benchmark of 5.8K pages across 82 languages and 9 tasks with unified metrics, enabling systematic cross-model comparison (\S\ref{sec:benchmark}).
    \item A systematic study showing DPO with rendering-derived ground truth outperforms supervised fine-tuning and closed-model distillation for cross-lingual transfer (\S\ref{sec:training}).
\end{itemize}

%% file: sec/2_related.tex
\section{Related Works}
\label{sec:related}

\begin{table}[t]
    \centering
    \scriptsize
    \renewcommand{\arraystretch}{1.2}
    \setlength{\tabcolsep}{2pt}
    \rowcolors{2}{gray!6}{white}
    \resizebox{\linewidth}{!}{
        \begin{tabular}{l c c c c c c}
            \toprule
            \textbf{Dataset} & \textbf{\# Samples} & \textbf{\# Lang.} & \textbf{Data Type} & \textbf{Annotation} & \textbf{\# Tasks} & \textbf{Model-Free} \\
            \midrule
            FUNSD         & 199        & 1  & Real      & Manual    & 1 & \cmark \\
            XFUND         & 1.3K      & 7  & Real      & Manual    & 2 & \cmark \\
            PubLayNet & 361K      & 1  & Real      & Rule-based & 3 & \cmark \\
            DocLayNet & 81K       & 4  & Real      & Manual & 4 & \cmark \\
            DocBank     & 500K       & 1  & Real      & Model-based & 3 & \xmark \\
            SynthTIGER & 10M     & 1  & Synthetic & Rule-based & 1 & \cmark \\
            WordScape      & 9.5M    & 136  & Real      & Colorization & 3 & \cmark \\
            DIT700K           & 700K    & 1  & Real      & Layout detector & 3 & \xmark \\
            \midrule
            \rowcolor{blue!12}
            \textbf{DocAtlas (ours)}            & \textbf{360K}      & \textbf{82} & \textbf{Real+Synth(15\%)} & \textbf{Diff. Rendering} & \textbf{9} & \textbf{\cmark} \\
            \bottomrule
        \end{tabular}
    }
    \caption{\textbf{Comparison of OCR training datasets.} DocAtlas achieves 82 languages (11.7× more than XFUND) and 9 tasks (3× more than competitors) through rendering-based annotation that eliminates model dependency.}
    \label{tab:dataset_comparison}
    \vspace{-0.5em}
\end{table}

\paragraph{OCR Dataset Construction.}
Traditional datasets rely on manual annotation (FUNSD~\cite{jaume2019funsd}, XFUND~\cite{xu2022xfund}), limiting scalability. Synthetic pipelines (SynthTIGER~\cite{yim2021synthtiger}, DocCreator~\cite{journet2017doccreator}, Donut~\cite{donut}) use forward generation with text at predefined positions; annotations are trivially correct but cannot capture real document complexity such as nested tables or authentic formatting. Large-scale model-based efforts (PubLayNet~\cite{zhong2019publaynet}, DocBank~\cite{li2020docbank}, DIT700K~\cite{dit700}) automate annotation using pretrained layout detectors, creating circular dependency where quality is bounded by existing model performance.

Rendering-based pipelines offer a middle path. WordScape~\cite{weber2023wordscape} recovers layout from Common Crawl Word documents via colorization, but relies on LibreOffice conversion (introducing rendering drift from font substitution and text reflow), extracts text independently of bounding boxes without geometric alignment guarantees, treats charts as opaque figures, and provides no RTL script coverage. We build upon this paradigm but treat rendering as a \emph{closed-form annotation function}: lossless MS Word rendering eliminates drift, pixel-wise differential subtraction disambiguates injected colors from pre-existing ones, and joint IoU-based text-geometry matching produces aligned DocTag annotations suitable for multi-task supervision. We detail these improvements in \S\ref{sec:pipeline_a}. Table~\ref{tab:dataset_comparison} summarizes the landscape.

\paragraph{Multilingual Model Training.}
Extending OCR to new languages without degrading original performance remains challenging due to catastrophic forgetting~\cite{luo2025empirical}. Parameter-efficient methods (e.g., LoRA~\cite{hu2022lora}) have been used for OCR adaptation~\cite{chung2025finetuning} with reduced memory. Component-level training (Dolphin~\cite{feng2025dolphin}, SmolDocling~\cite{smoldocling}) focuses supervision on specific elements rather than full pages. DPO~\cite{dpo} shows promise in preserving base capabilities. We systematically compare these strategies and additionally disentangle the effect of training algorithm from dataset quality by comparing DPO with rendering-derived ground truth against GPT-4o distillation (\S\ref{sec:training}).

\begin{table}[t!]
    \centering
    \scriptsize
    \renewcommand{\arraystretch}{1.12}
    \setlength{\tabcolsep}{1.7pt}
    \rowcolors{2}{gray!6}{white}
    \resizebox{\linewidth}{!}{
        \begin{tabular}{l c c c c c c c c c }
            \toprule
            \textbf{Benchmark} & \textbf{Year} & \textbf{Langs} &
            \textbf{\shortstack{E2E}} &
            \textbf{\shortstack{RO}} &
            \textbf{Text} & \textbf{Table} &
            \textbf{Formula} & \textbf{Chart}  \\
            \midrule
            PubTabNet~\cite{teds-score}         & 2019 & 1  & \xmark & \xmark & \xmark & \cmark & \xmark & \xmark   \\
            XFUND~\cite{xu2022xfund}             & 2022 & 7  & \xmark & \xmark & \cmark & \xmark & \xmark & \xmark  \\
            Nougat~\cite{blecher2023nougat}            & 2023 & 1  & \cmark & \xmark & \cmark & \cmark & \cmark & \xmark  \\
            READOC~\cite{li2025readoc}            & 2025 & 27 & \cmark & \cmark & \cmark & \cmark & \cmark & \xmark   \\
            OmniDocBench~\cite{omnidocbench}      & 2025 & 2  & \cmark & \cmark & \cmark & \cmark & \cmark & \xmark   \\
            \rowcolor{blue!8}
            DocAtlas (ours)   & 2025 & 82 & \cmark & \cmark & \cmark & \cmark & \cmark & \cmark  \\
            \bottomrule
        \end{tabular}
    }
    \caption{\textbf{Document parsing benchmarks.} DocAtlas offers 3$\times$ more languages and supports all major elements. E2E: End-to-End, RO: Reading Order.}
    \label{tab:benchmark_comparison}
    \vspace{-0.6em}
\end{table}
\paragraph{Evaluation Benchmarks.}
Existing benchmarks vary in scope: PubTabNet~\cite{teds-score} focuses on tables, XFUND~\cite{xu2022xfund} covers 7 languages, while recent efforts (READOC~\cite{li2025readoc}, OmniDocBench~\cite{omnidocbench}, Docling-Eval~\cite{auer2024docling}) expand task coverage but remain limited in language diversity (27, 2, and 4 languages). Evaluation fragmentation prevents direct comparison across systems. Table~\ref{tab:benchmark_comparison} shows no existing benchmark simultaneously covers diverse languages and comprehensive document elements for parsing evaluation.

%% file: sec/3_methods.tex
\section{Methods}
\label{sec:methods}

To construct large-scale multilingual OCR supervision without model dependency, we developed complementary pipelines (Figures~\ref{fig:generation_pipeline},\ref{fig:rtl_pipeline}). The first processes native Word documents from Common Crawl, while the second synthesizes right-to-left documents to fill gaps in script coverage.

\subsection{Pipeline A: Native Word Documents}
\label{sec:pipeline_a}

Inspired by~\cite{weber2023wordscape}, we begin by parsing \texttt{.wat} metadata from Common Crawl to extract candidate \texttt{.doc}/\texttt{.docx} URLs. Canonicalization-based deduplication is applied within each snapshot, and a RocksDB \cite{rocksdb2024} key–value store ensures cross-snapshot deduplication, filtering out $60$-$80\%$ of redundant URLs. Once URLs are extracted, the corresponding files are downloaded along with provenance metadata. During this stage, unsafe documents, those containing macros, embedded objects, or encryption, are automatically discarded, as are oversized or zip-bomb-like archives. SHA-256 hashing is applied to ensure content-level deduplication and integrity, and any file that fails to open or exhibits corrupted structure is logged and removed.

\begin{figure*}
  \centering
  \includegraphics[width=\linewidth]{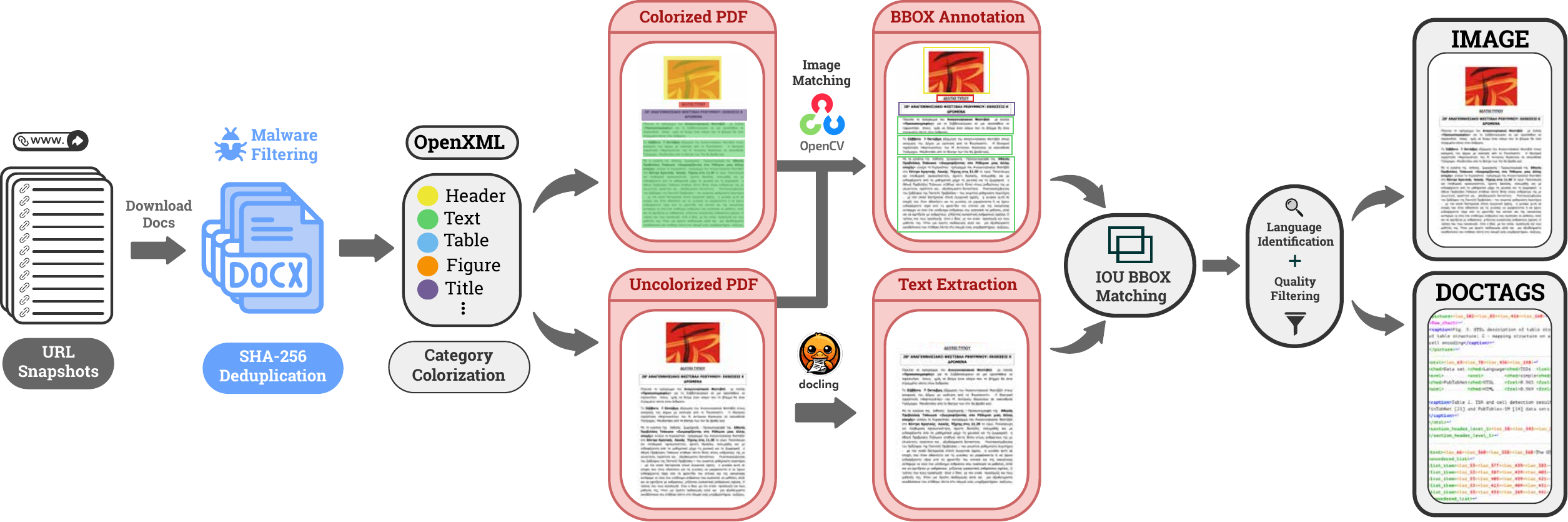}
  \vspace{-1.5em}
  \caption{\textbf{End-to-end data pipelines.} We implement two pipelines: a high-fidelity pipeline for native DOCX documents and a synthetic RTL pipeline for underrepresented scripts. The native pipeline extracts, filters, colorizes, and annotates Word files, while the RTL pipeline converts structured inputs (EPUB, HTML, XML) into precisely annotated PDF documents using LaTeX synthesis.}
  \label{fig:generation_pipeline}
  \vspace{-0.5em}
\end{figure*}

After acquiring a clean set of documents, we recover structure directly from OpenXML markup. Components are identified from native tags and built-in styles (e.g., tables, figures) and further refined using heuristic cues such as font size and list patterns. To distinguish component types, we inject color codes via Word styling attributes, then render both colorized and uncolorized versions to PDF (Figure~\ref{fig:generation_pipeline}). Subtracting these two renderings pixel-wise yields precise per-category bounding boxes through OpenCV contour analysis \cite{opencv_library}, producing high-quality, model-free annotations from rendering differences alone.

With structure recovered, we align textual content to its geometric layout. Text is extracted at the document level from OpenXML and at the page level using the Docling~\cite{docling} rule-based PDF parser (analogous to PyMuPDF, not a neural model). Word-level boxes are then matched to component regions using intersection-over-union (IoU) containment. When components overlap, such as text boxes drawn over images, we resolve conflicts by prioritizing the component with higher style-based confidence, ensuring consistent structural mapping across complex layouts.

\paragraph{Quality Filtering and Perplexity Analysis.}
To maintain high multilingual quality, we apply a two-stage filtering process. First, we predict document language using fastText~\cite{fasttext} and compute perplexity via language-specific 5-gram Kneser–Ney models~\cite{ccnet}, thresholding at $\tau = 120$ to retain over 94\% of high-quality data while filtering out 38\% of low-confidence pages. Second, we compute an \textit{annotation reliability score} based on the proportion of characters tagged via native XML signals rather than heuristics, excluding pages below 0.6 along with those exhibiting anomalous visual signals, resulting in roughly 15\% removal following~\cite{weber2023wordscape}. Additional filtering details and per-language perplexity distributions are provided in Appendix~\ref{appendix:quality-filtering}. Real-world documents with complex backgrounds, nested tables, rotated text, or embedded advertisements are automatically detected and excluded to avoid propagating noisy supervision, preserving annotation precision at the cost of roughly 15\% volume reduction.

Finally, we serialize all pages into the DocTag~\cite{smoldocling} format, a unified XML-like schema encoding component type, geometry, and text content as shown in Figure~\ref{fig:generation_pipeline}. Unlike HTML, which omits layout geometry, or Markdown, which collapses hierarchy, DocTag preserves both structure and semantics, enabling multi-task supervision for layout detection, reading order, and content extraction. Each page becomes a flat tag sequence (e.g., \textbf{\textcolor{mycmd}{\texttt{<text>}}}, \textbf{\textcolor{mycmd}{\texttt{<section\_header>}}}, \textbf{\textcolor{mycmd}{\texttt{<table>}}}) with corresponding bounding boxes. To support flexible downstream use, we provide multiple output variants, including JSON, HTML, Markdown, and visual overlays.

Beyond basic annotations, we enrich each page with semantic metadata. Captions (\textbf{\textcolor{mycmd}{\texttt{<figure\_caption>}}}, \textbf{\textcolor{mycmd}{\texttt{<table\_caption>}}}) are identified through XML style cues and linguistic prefixes, then linked to nearest visual components via vertical adjacency. Figures are classified into categories (\emph{natural image}, \emph{logo}, \emph{QR code}, \emph{chart}, \emph{graph}) via Docling~\cite{docling}, equations normalized to \LaTeX{}, and page-level attributes (column count, watermark, background type) are inferred by Qwen3-VL~\cite{qwen3}. These two model-based steps provide \emph{optional metadata enrichment only}, all core DocTag annotations are produced entirely through differential rendering and OpenXML parsing without learned models (Table~\ref{tab:model_dependency}).

\begin{figure}[t]
    \centering
    \includegraphics[width=\linewidth]{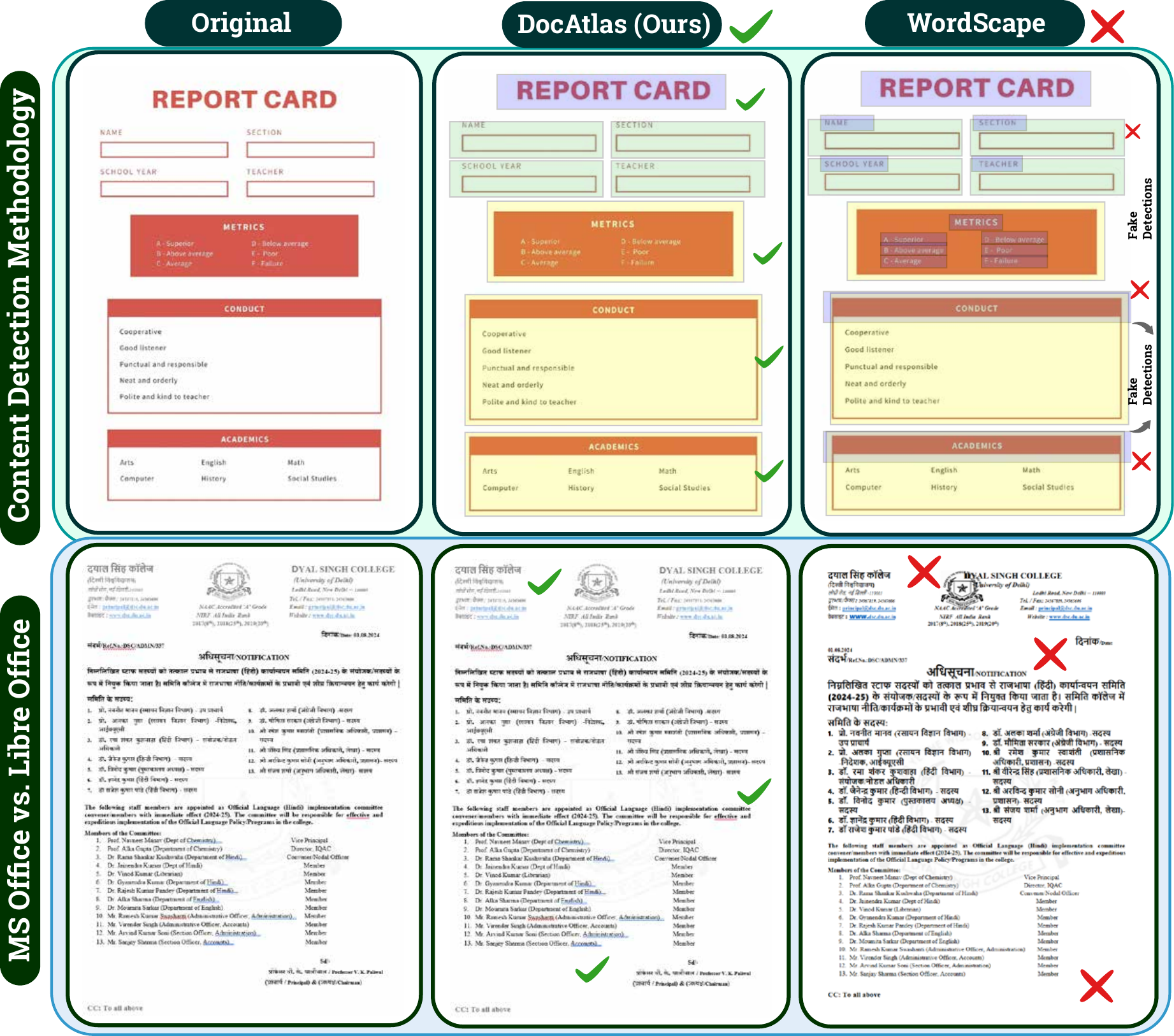}
    \vspace{-1.0em}
    \caption{\textbf{Comparison with WordScape.} (Top) MS Word rendering preserves layout fidelity; LibreOffice introduces drift. (Bottom) Differential rendering eliminates false detections from pre-existing colors.}
    \label{fig:wordscape_comparison}
    \vspace{-0.5em}
\end{figure}

\paragraph{Comparison with WordScape.}
Although our pipeline builds upon the Common Crawl extraction strategy of WordScape~\cite{weber2023wordscape}, the annotation methodology differs in three critical respects (Figure~\ref{fig:wordscape_comparison}): (1)~pixel-wise differential rendering disambiguates injected color codes from pre-existing document colors, which single-pass colorization cannot; (2)~we render through MS Word rather than LibreOffice, eliminating stochastic drift from font substitution and text reflow; and (3)~word-level IoU matching jointly encodes text, geometry, and component type, replacing fragmented JSON extraction with no alignment guarantees. Together, these enable multi-task supervision (layout + reading order + content extraction) rather than coarse layout detection alone.

\subsection{Pipeline B: Synthetic RTL Pipeline}
\label{sec:pipeline_b}

\begin{figure}[t]
    \centering
    \includegraphics[width=\linewidth, trim=15 10 15 10, clip]{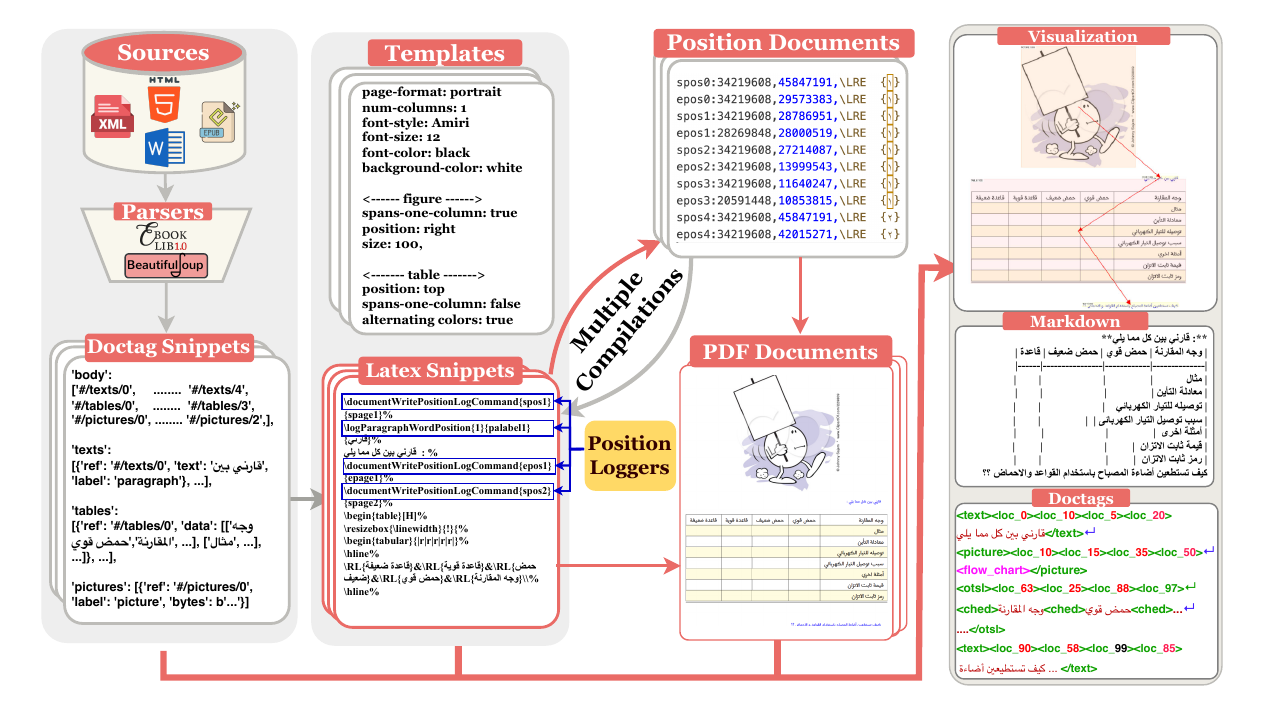}
    \vspace{-1.5em}
    \caption{\textbf{Overview of the DocAtlas synthetic data generation pipeline.}
    Structured inputs (HTML, XML, DOCX, EPUB) are parsed into \texttt{DocTag} snippets
    and rendered via \LaTeX{} templates with positional logging.
    Through multiple compilations, the system produces aligned \texttt{PDF} documents
    and precise element-level annotations (\texttt{DocTag}, Markdown, and visual overlays).}
    \label{fig:rtl_pipeline}
    \vspace{-0.5em}
\end{figure}

While the native pipeline effectively covers left-to-right scripts, right-to-left (RTL) languages remain underrepresented due to parsing failures in existing PDF tools. To close this gap, we introduce a synthetic generation pipeline that produces near-perfectly annotated RTL documents through \LaTeX{}-based rendering (Figure~\ref{fig:rtl_pipeline}). Structured inputs (EPUB, HTML, XML) are parsed into a standardized Docling JSON schema, where each content element is tagged and assigned provisional bounding boxes. Document synthesis proceeds through 205 LuaTeX-based templates covering Arabic, Hebrew, Urdu, and Persian, governing typography, layout, and bidirectional text control:
\begin{equation}
\text{Docling} + \text{Template} \xrightarrow{\text{LaTeX}} \text{PDF} + \text{.pos}.
\end{equation}
Custom \LaTeX{} commands log positional metadata during three compilation passes (initial layout, position logging, final rendering), enabling exact bounding-box recovery for all elements. The resulting output pairs a rendered PDF with Docling~\cite{docling} JSON containing element-level bounding boxes, text content, and structural labels. The pipeline generates 52K pages across 4 RTL languages with near-perfect annotation precision; implementation details including coordinate transformations, bidirectional markers, and chart synthesis are provided in Appendix~\ref{sec:pipeline_b_details}.

\subsection{Benchmark}\label{sec:benchmark}

We assembled a multilingual benchmark balancing diversity, difficulty, and representativeness. Samples are drawn from the training corpus and targeted additions emphasizing rare structures (charts, formulas, multi-task layouts). Following~\cite{omnidocbench}, pages are embedded with ResNet-50~\cite{resnet} features, clustered via FAISS~\cite{faiss}, and stratified by difficulty into equal easy/medium/hard splits, yielding up to 100 pages per language across 82 languages (5,575 samples). We curate 144 challenging formula samples and generate multilingual chart data across 15 languages using a VLM-seeded pipeline with expert verification ($\kappa$=0.89; details in Appendix~\ref{appendix:benchmark}). Evaluation covers end-to-end page-to-Markdown/DocTag conversion, measured via text edit distance, TEDS~\cite{teds-score} for tables, formula transcription accuracy, and reading order fidelity. Additional subtasks, chart$\to$HTML, formula$\to$\LaTeX, and table$\to$HTML, extend evaluation to 9 tasks.

\subsection{Multilingual Training Enrichment}\label{sec:training}

We investigate three training strategies for extending OCR models to new languages while mitigating catastrophic forgetting: (i)~full-page SFT on page$\to$DocTag/Markdown pairs, (ii)~component-level SFT on cropped elements (paragraphs, tables, charts, formulas), and (iii)~DPO~\cite{dpo}, which preserves base-language behavior by preferring rendering-derived ground truth over base model predictions. We further vary the subset of trainable parameters (QKV, MLP, or full model) to evaluate the gain-forgetting trade-off.

%% file: sec/4_experiments.tex
\section{Analysis \& Experiments}\label{sec:experiments}

\begin{table*}[t!]
  \centering
  \resizebox{\textwidth}{!}{
    \begin{tabular}{c|l|c|l|c|c|c|c|c}
    \toprule
    \textbf{Method Type} & \textbf{Methods} & \textbf{Params} & 
    \textbf{{Text\textsuperscript{Edit}}}$\downarrow$ & \textbf{{Table\textsuperscript{TEDS}}}$\uparrow$ & \textbf{{Formula\textsuperscript{Edit}}}$\downarrow$ & 
    \textbf{{Read Order\textsuperscript{Edit}}}$\downarrow$ & \textbf{{Overall\textsuperscript{Edit}}}$\downarrow$ \\
    
     \midrule
     \multirow{3}{*}{\makecell{\textbf{General VLMs}}} 
     & Gemini-2.0-Pro~\cite{gemini25} & - &  0.090 & 68.50 & 0.356  & 0.050 & 79.75 \\
     & GPT4o~\cite{gpt4o} & - &  0.117 & 62.26 & 0.425 & 0.065 & 75.30 \\
     & Qwen3-VL~\cite{qwen3} & 3B & 0.081 & 51.86 & 0.420 & 0.089 & 71.87 \\
     & Qwen2.5-VL~\cite{qwen25} & 2B &  0.174 & 50.59 & 0.453 & 0.117 & 66.59 \\
     & InternVL3.5~\cite{internvl35} & 2B &  0.095 & 70.80 & 0.543 & 0.060 & 77.20 \\
     \midrule
     \multirow{2}{*}{\makecell{\textbf{Expert VLMs}}} 
     & DotsOCR~\cite{dotsocr2025} & 3B &  0.068 & 65.40 & 0.321 & \textbf{0.037} & 79.29 \\
     & PaddleOCR-VL~\cite{paddleocrvl} & 1B &  0.078 & 73.90 & 0.241 & 0.052 & 80.10 \\
     & DeepseekOCR~\cite{deepseekocr2025} & 3B &  0.082 & 71.54 & 0.242 & 0.053 & 81.66 \\
     & MonkeyOCR-pro~\cite{li2025monkeyocr} & 1.2B & 0.095 & 72.80 & 0.295 & 0.065 & 78.25 \\
     & Dolphin~\cite{feng2025dolphin} & 400M & 0.160 & 58.30 & 0.465 & 0.066 & 71.17 \\
     & Nanonets-OCR-s~\cite{nanonetsocr} & 4B & 0.088 & 71.90 & 0.518 & 0.059 & 81.53 \\
     & Nanonets-OCR2~\cite{nanonetsocr2} & 3B & 0.088 & 66.24 & 0.471 & 0.060 & 78.70 \\
     & Chandra~\cite{chandra} & 9B & 0.071 & 69.79 & 0.262 & 0.042 & 81.33 \\
     & MinerU2.5~\cite{mineruv25} & 1.2B & 0.267 & \textbf{72.79} & 0.273 & 0.096 & 73.07 \\
     \rowcolor{blue!8}
     & DocAtlas-Deepseek (Ours) & 3B & \textbf{0.055} & 72.24 & \textbf{0.237} & 0.049 & \textbf{83.37} \\
     
    \bottomrule
    \end{tabular}%
  }
  \caption{
    \textbf{Quantitative comparison across OCR systems on our multilingual benchmark.} 
    We report character-level text recognition accuracy (\textsc{Text\textsubscript{Edit}↓}), table structure accuracy using TEDS~\cite{teds-score} (\textsc{Table\textsubscript{TEDS}↑}), formula transcription accuracy (\textsc{Formula\textsubscript{Edit}↓}), and reading order fidelity (\textsc{Read Order\textsubscript{Edit}↓}). 
    \textsc{Overall\textsubscript{Edit}↓} represents the average of text and table scores (after converting text edit distance to accuracy). 
    DocAtlas achieves the highest overall performance, outperforming prior expert and general-purpose VLMs baselines.
    }
  \label{tab:results}
  \vspace{-0.5em}
\end{table*}

\paragraph{Dataset Statistics and Quality Control.}
We sourced 1.9M documents spanning 5.48M pages across 136 languages from Common Crawl under permissive licenses, with automated PII detection removing 5.15\% of documents. Our native pipeline (Pipeline A) sustains 100k+ pages/day on a single CPU, while the synthetic RTL pipeline (Pipeline B) generates 195k pages at 183 pages/minute. Three document classes require targeted filtration, scanned PDFs (8.2\%, excluded), rendering drift from missing fonts ($<$0.3\%, mitigated via tolerance-aware contour matching), and malformed OpenXML (repaired via schema validation), ensuring 98.9\% of retained documents maintain $>$95\% annotation accuracy. After quality filtering and difficulty-aware sampling, the final corpus comprises 360k training pages across 82 languages, 31 structural element types, and 25+ content domains. Comprehensive details on collection, licensing, efficiency, and component distributions are provided in Appendices~\ref{app:data_generation}-\ref{sec:domain_diversity}.

\paragraph{Model Selection and Evaluation Methodology.}
We evaluated 14 models spanning general VLMs~\cite{gpt4o,qwen3,qwen25,gemini25,internvl35} (multilingual baselines without layout training), expert document models~\cite{smoldocling,granitedocling,dotsocr2025} (compact layout-grounded parsing), and OCR-specific systems~\cite{deepseekocr2025,nanonetsocr,nanonetsocr2,feng2025dolphin,paddleocrvl,li2025monkeyocr} (cross-script supervision with structural output), enabling controlled analysis across architecture, scale, and training paradigms. We inference each model for Markdown outputs, then apply a three-stage pipeline~\cite{omnidocbench}: extraction (LaTeX/HTML tables, formulas, paragraphs with inline LaTeX→Unicode conversion), fuzzy Adjacency Search Match~\cite{omnidocbench} using Normalized Edit Distance (direct matching for high-confidence pairs, iterative merging for partials), and metric computation across full-page parsing, individual tasks (text, table, formula, reading order), and condition-specific attributes (layout type, watermark, merged cells), ignoring headers/footers/captions. Detailed metrics are in Appendix~\ref{sec:metrics}. Additionally, training and data generation setups are in Appendix~\ref{appendix:setup} and layout robustness in Appendix~\ref{appendix:layout-robust}.

%% file: sec/5_results.tex
\section{Results}
\label{sec:results}

\subsection{Leaderboard Comparison}

Our benchmark evaluation reveals critical performance patterns across multilingual document understanding. In Table~\ref{tab:results}, DocAtlas-Deepseek achieves state-of-the-art performance (83.37\% overall), with DeepseekOCR following closely at 81.66\% despite being a compact 3B model, demonstrating remarkable efficiency in balancing model size with accuracy. Notably, text recognition substantially outperforms structured content extraction across all systems: text edit distances average 0.068--0.095 for top models, while table TEDS scores plateau at 71--73\%, highlighting that spatial reasoning over complex layouts remains a fundamental challenge. We identify 88,036 errors across 12 categories, with four dominant types: table spanning errors (15.7\%), formatting (14.6\%), character encoding (13.2\%), and content omission (13.2\%). These affect table structure, text styling, Unicode normalization, and list/hyphen handling.

\begin{figure}[h]
    \centering
    \includegraphics[width=\linewidth]{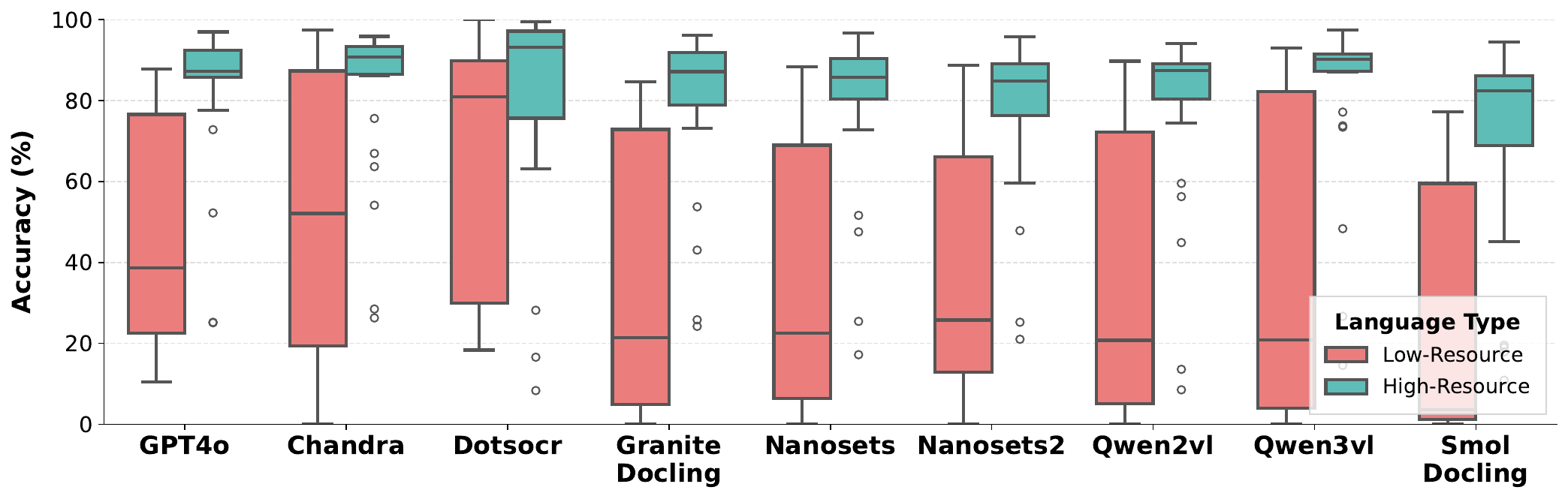}
    \vspace{-1.5em}
    \caption{\textbf{Accuracy distribution across high- and low-resource languages.}}
    \label{fig:low-vs-high-resource}
\end{figure}
Figure~\ref{fig:low-vs-high-resource} exposes a stark resource divide: high-resource languages maintain consistent 80-95\% accuracy with narrow variance, while low-resource scripts exhibit 20-85\% accuracy ranges with median performance often below 40\%, underscoring how training data availability dictates multilingual robustness more than architectural sophistication.
\begin{figure}[h]
    \centering
    \includegraphics[width=\linewidth]{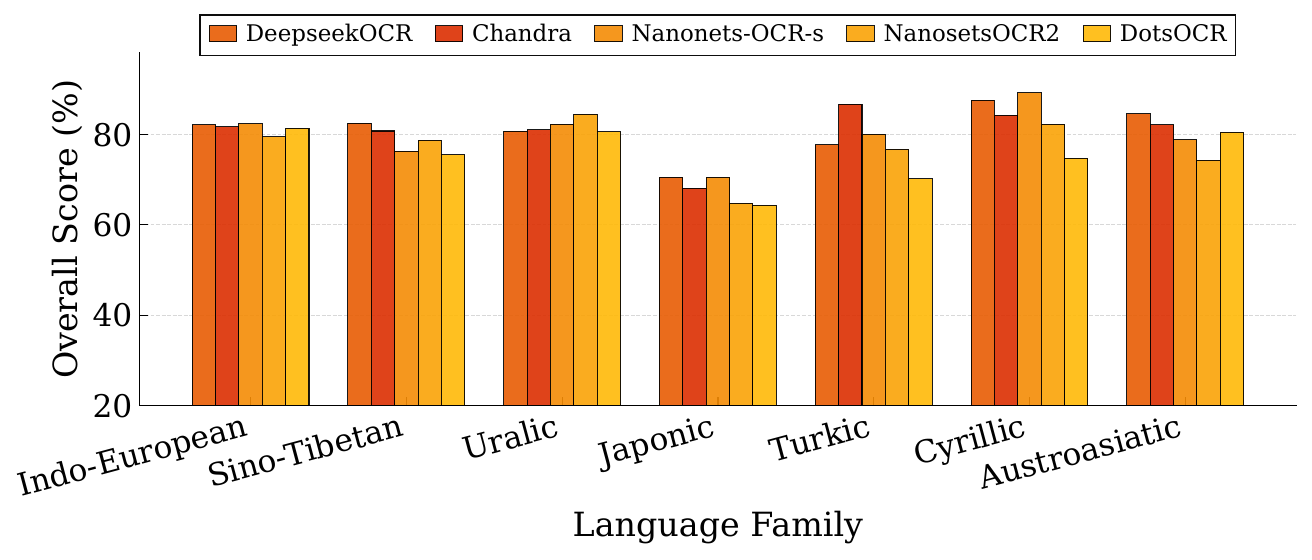}
    \vspace{-1.5em}
    \caption{\textbf{OCR accuracy across language families.} Scores (brighter is better) show average performance across 14 models and 7 families. Top models (e.g., DeepseekOCR, Chandra) are consistent, while others degrade on low-resource scripts.}

    \label{fig:lang-families}
    \vspace{-0.5em}
\end{figure}
Cross-linguistic and domain-specific analysis reveals systematic biases in current OCR training paradigms. Language family performance (Figure~\ref{fig:lang-families}) shows Indo-European and Cyrillic scripts achieving 80-87\% accuracy, contrasting sharply with Japonic (26.9-70.5\%) and Austroasiatic families where even top models struggle, suggesting that morphological complexity and logographic systems expose fundamental gaps in visual feature learning.

\begin{figure}[h]
    \centering
    \includegraphics[width=\linewidth]{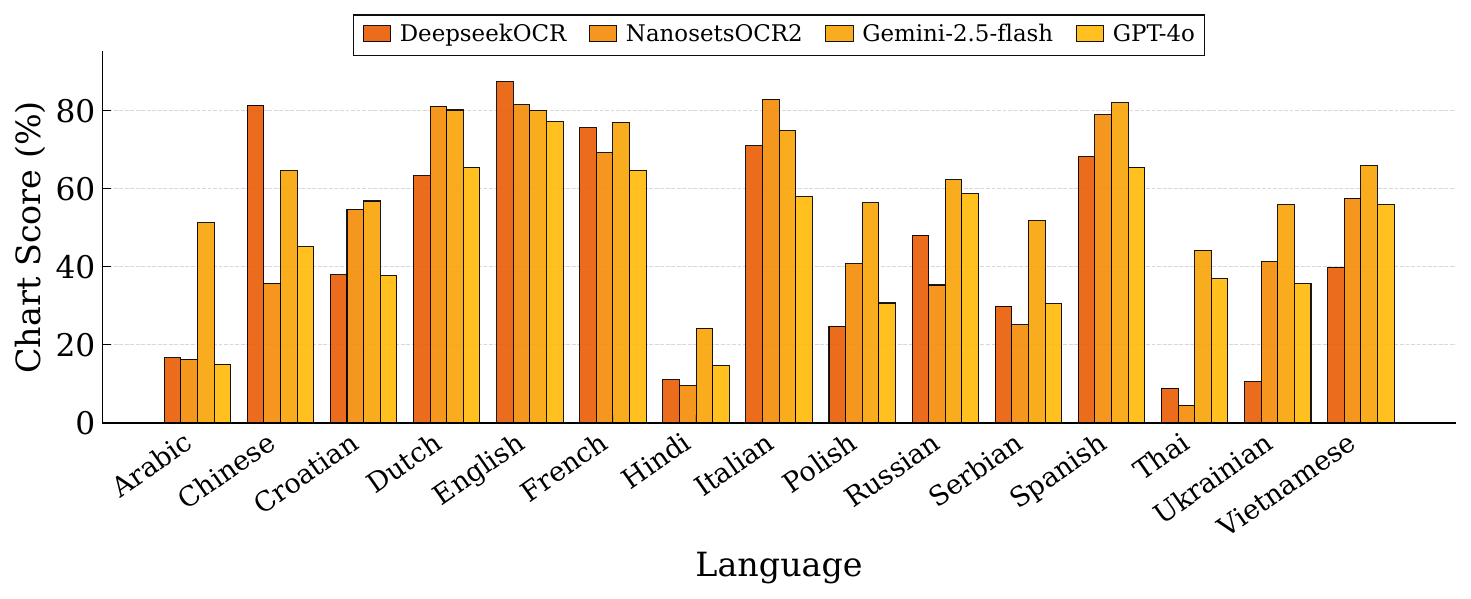}
    \vspace{-1.5em}
    \caption{\textbf{Chart extraction accuracy across 15 languages.} Gemini-2.5-Flash achieves the highest average.}
    \label{fig:chart_heatmap}
    \vspace{-0.5em}
\end{figure}

\begin{table}[ht]
    \centering
    
    \rowcolors{2}{white}{gray!10}
    \footnotesize  
    \scalebox{0.8}{
  \begin{tabular}{lccc}
      \toprule
      \textbf{Model} & \textbf{Bar} & \textbf{Line} & \textbf{Pie} \\
      \midrule
      DeepseekOCR~\cite{deepseekocr2025}      & 0.195 & 0.649 & 0.522 \\
      SmolDocling~\cite{smoldocling}          & 0.127 & 0.038 & 0.337 \\
      NanosetsOCR2~\cite{nanonetsocr}         & 0.397 & 0.603 & 0.446 \\
      Gemini-2.5-flash~\cite{gemini25}        & \textbf{0.471} & \textbf{0.662} & \textbf{0.673} \\
      GPT-4o~\cite{gpt4o}                     & 0.280 & 0.405 & 0.566 \\
      \bottomrule
      \end{tabular}
      }
      \caption{\textbf{Chart type accuracy.} Mean OCR scores for each model across 3 chart types. Gemini-2.5-flash~\cite{gemini25} performs best overall.}
    \label{tab:model_chart_scores}
    \vspace{-0.5em}
\end{table}
\paragraph{Multilingual Chart Extraction}
Chart extraction reveals a critical divide between specialized OCR systems and general-purpose vision-language models. As shown in Figure~\ref{fig:chart_heatmap}, Gemini-2.5-Flash achieves the highest average performance (61.82\%) with cross-lingual consistency, while expert OCR models exhibit severe language-specific degradation, DeepseekOCR scores 87\% on English but collapses to 8-17\% on Thai, Arabic, and Italian. Table~\ref{tab:model_chart_scores} shows this gap persists across chart types, with multimodal models consistently outperforming document-specific architectures (e.g., SmolDocling's near-zero line chart accuracy of 0.038). These findings indicate that robust multilingual chart parsing requires visual reasoning beyond text extraction, motivating our inclusion of synthetic chart data across typologically diverse scripts.

\begin{table}[h!]
\centering
\resizebox{\linewidth}{!}{
\begin{tabular}{l@{\hspace{8pt}}c@{\hspace{4pt}}c@{\hspace{12pt}}c@{\hspace{4pt}}c}
\toprule
\multirow{2}{*}{\textbf{Method}} & 
\multicolumn{2}{c}{\textbf{Text Edit Distance ($\downarrow$)}} &
\multicolumn{2}{c}{\textbf{Table TEDS ($\uparrow$)}} \\
\cmidrule(lr){2-3} \cmidrule(lr){4-5}
& \textbf{New Lang.} & \textbf{Base Lang.} & \textbf{New Lang.} & \textbf{Base Lang.} \\
\midrule
Full-SFT (All modules) & $-0.038$ & $+0.118$ & $+13.6$ & $-12.1$ \\
\midrule
\multicolumn{5}{l}{\rotatebox{0}{\textit{LoRA Variants:}}} \\
\rowcolor{gray!10}
\quad All layers       & $-0.025$ & $+0.069$  & $+8.9$  & $-5.6$  \\
\quad MLP only         & $-0.031$ & $+0.042$  & $+10.1$ & $-3.9$  \\
\rowcolor{gray!10}
\quad MLP Gate\&Down   & $-0.034$ & $+0.028$  & $+10.8$ & $-2.7$  \\
\quad All QKV          & $-0.027$ & $+0.053$  & $+9.7$  & $-1.9$  \\
\rowcolor{blue!8}
\quad \textbf{QKV only}& \textbf{$-0.021$} & \textbf{$-0.011$} & \textbf{$+9.2$} & \textbf{$+1.3$} \\
\bottomrule
\end{tabular}
}
\caption{
\textbf{Multilingual adaptation: gain vs. forgetting trade-off.}
All changes measured relative to baseline (Text Edit: 0.082, Table TEDS: 71.5).
LoRA (QKV only) achieves optimal balance.
}
\label{tab:lora_analysis}
\vspace{-0.5em}
\end{table}

\subsection{Training Strategy Analysis}

\textbf{Layer-wise adaptation trade-offs.} Table~\ref{tab:lora_analysis} shows that full SFT achieves the largest new-language gains (+13.6 TEDS) but severely degrades base-language performance ($-$12.1 TEDS). Among LoRA~\cite{hu2022lora} variants, QKV-only training achieves optimal balance (confirmed by~\cite{zhu2025teach}): $-$0.021 edit distance on new languages while improving base performance ($-$0.011). We attribute this to a functional asymmetry: QKV governs \emph{where} the model attends, while MLP layers shape output token distributions. Tuning MLP shifts distributions toward task-specific tokens, causing forgetting; QKV-only adaptation learns cross-lingual attention routing without biasing outputs.
\begin{table}[h!]
\centering

\resizebox{\linewidth}{!}{
\begin{tabular}{l@{\hspace{6pt}}rr@{\hspace{10pt}}rr@{\hspace{10pt}}rr@{\hspace{10pt}}rr}
\toprule
\textbf{Model} & 
\multicolumn{2}{c@{\hspace{10pt}}}{\textbf{Baseline}} & 
\multicolumn{2}{c@{\hspace{10pt}}}{\textbf{Full-Page SFT}} & 
\multicolumn{2}{c@{\hspace{10pt}}}{\textbf{Component SFT}} & 
\multicolumn{2}{c}{\textbf{DPO}} \\
\cmidrule(lr){2-3} \cmidrule(lr){4-5} \cmidrule(lr){6-7} \cmidrule(lr){8-9}
& \small In & \small Out & 
\small In & \small Out & 
\small In & \small Out & 
\small In & \small Out \\
\midrule
\rowcolor{gray!8}
Qwen2.5-VL & 
\textcolor{gray}{66.6} & \textcolor{gray}{61.9} & 
70.5 \textcolor{green!50!black}{\small (+3.9)} & 53.6 \textcolor{red}{\small (-8.2)} &
71.2 \textcolor{green!50!black}{\small (+4.6)} & 44.5 \textcolor{red}{\small (-17.4)} & 
\cellcolor{blue!12}\textbf{68.1} \textcolor{green!50!black}{\small \textbf{(+1.5)}} & 
\cellcolor{blue!12}\textbf{63.3} \textcolor{green!50!black}{\small \textbf{(+1.4)}} \\

Nanonets-OCR & 
\textcolor{gray}{81.5} & \textcolor{gray}{75.7} & 
86.3 \textcolor{green!50!black}{\small (+4.8)} & 65.7 \textcolor{red}{\small (-10.1)} &
87.1 \textcolor{green!50!black}{\small (+5.6)} & 54.5 \textcolor{red}{\small (-21.2)} & 
\cellcolor{blue!12}\textbf{83.4} \textcolor{green!50!black}{\small \textbf{(+1.9)}} & 
\cellcolor{blue!12}\textbf{77.6} \textcolor{green!50!black}{\small \textbf{(+1.8)}} \\

\rowcolor{gray!8}
DotsOCR & 
\textcolor{gray}{79.3} & \textcolor{gray}{73.7} & 
83.9 \textcolor{green!50!black}{\small (+4.6)} & 63.9 \textcolor{red}{\small (-9.8)} &
84.7 \textcolor{green!50!black}{\small (+5.4)} & 53.0 \textcolor{red}{\small (-20.7)} & 
\cellcolor{blue!12}\textbf{81.1} \textcolor{green!50!black}{\small \textbf{(+1.8)}} & 
\cellcolor{blue!12}\textbf{75.4} \textcolor{green!50!black}{\small \textbf{(+1.8)}} \\

DeepseekOCR & 
\textcolor{gray}{81.7} & \textcolor{gray}{75.9} & 
86.4 \textcolor{green!50!black}{\small (+4.8)} & 65.8 \textcolor{red}{\small (-10.1)} &
87.3 \textcolor{green!50!black}{\small (+5.6)} & 54.6 \textcolor{red}{\small (-21.3)} & 
\cellcolor{blue!12}\textbf{83.6} \textcolor{green!50!black}{\small \textbf{(+1.9)}} & 
\cellcolor{blue!12}\textbf{77.7} \textcolor{green!50!black}{\small \textbf{(+1.8)}} \\
\bottomrule
\end{tabular}
}
\caption{\textbf{Multilingual OCR adaptation strategies.} Performance on in-domain (In: trained languages) and out-of-domain (Out: unseen languages) test sets.}
\label{tab:training_strategies}
\vspace{-0.5em}
\end{table}
\textbf{DPO enables stable cross-lingual transfer.} Table~\ref{tab:training_strategies} demonstrates that DPO~\cite{dpo} fundamentally breaks the adaptation-forgetting trade-off observed in supervised methods. While Full-Page SFT and Component-level training exhibit an inverse relationship between specialization and retention, DPO uniquely improves both metrics simultaneously. Component-level training achieves the highest in-domain gains but suffers catastrophic forgetting (up to -21.3\%), suggesting that isolated document elements create brittle representations. This pattern holds across all architectures, indicating that using base model predictions as negative examples provides a fundamental mechanism for capability preservation.

\begin{table}[h!]
\centering

\vspace{-0.3em}
\resizebox{\linewidth}{!}{
\begin{tabular}{lcc}
\toprule
\textbf{DPO Positive Signal} & \textbf{In-Domain} & \textbf{Out-Domain} \\
\midrule
Baseline (no DPO) & 81.7 & 75.9 \\
GPT-4o distillation & 82.1 \textcolor{green!50!black}{\small (+0.4)} & 75.2 \textcolor{red}{\small ($-$0.7)} \\
\rowcolor{blue!8}
\textbf{DocAtlas GT (ours)} & \textbf{83.6} \textcolor{green!50!black}{\small \textbf{(+1.9)}} & \textbf{77.7} \textcolor{green!50!black}{\small \textbf{(+1.8)}} \\
\bottomrule
\end{tabular}
}
\caption{
\textbf{Disentangling DPO from dataset quality.} DPO on DeepseekOCR using different positive signals. GPT-4o distillation hurts out-of-domain transfer due to biases on low-resource scripts; rendering-derived ground truth provides unbiased supervision.
}
\label{tab:dpo_ablation}
\vspace{-0.5em}
\end{table}

\paragraph{Dataset quality drives DPO gains.} To disentangle method from training signal, we compare DPO with rendering-derived ground truth against DPO with GPT-4o outputs as positives (Table~\ref{tab:dpo_ablation}). GPT-4o distillation yields marginal in-domain gains (+0.4) but degrades out-of-domain transfer ($-$0.7), as GPT-4o introduces systematic biases on low-resource scripts, hallucinated diacritics, RTL column misordering, that propagate through distillation. This validates that the annotation pipeline, not DPO alone, drives cross-lingual improvement.

\paragraph{Generalization to out-of-distribution documents.} DPO-trained models also generalize beyond digital-native sources: on DocPTBench and OmniDocBench, both English-dominated, photographed/scanned benchmarks unseen during training, DocAtlas-DeepSeek reduces edit distance from 22.1\%$\to$20.7\% and 0.137$\to$0.122 over baseline DeepseekOCR, suggesting that cross-lingual attention routing learned via DPO transfers beyond the training domain.

\begin{figure}[h]
    \centering
    \includegraphics[width=\linewidth]{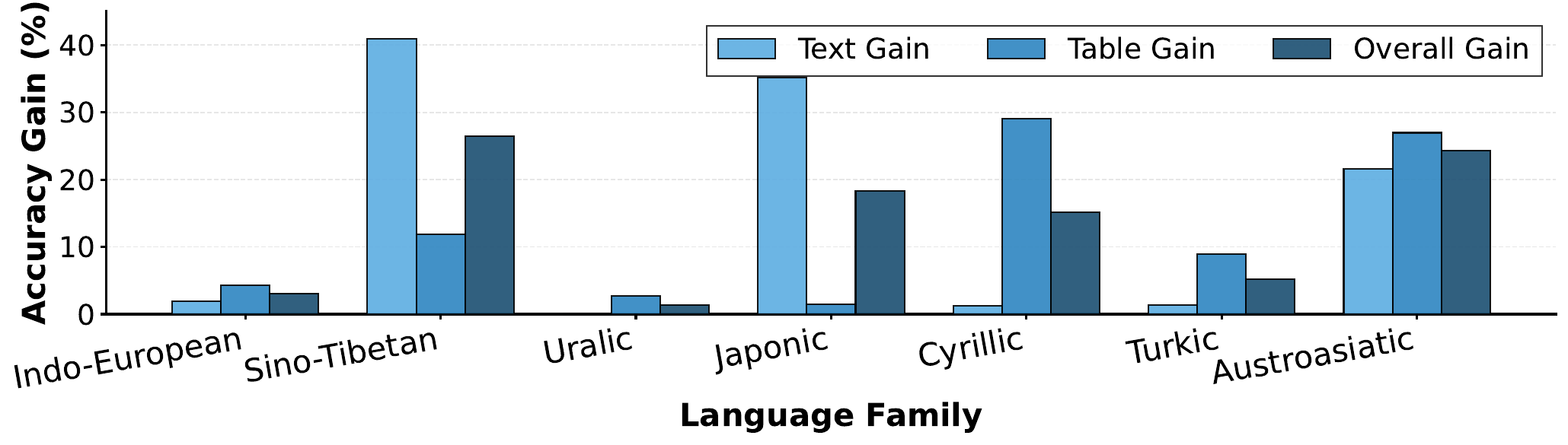}
    \vspace{-1.5em}
    \caption{\textbf{DPO gains across language families.}}

    \label{fig:lang-famaily-gain}
    \vspace{-0.5em}
\end{figure}
\paragraph{Language family gains reveal typological patterns.} Figure~\ref{fig:lang-famaily-gain} shows that DPO training benefits vary significantly across language families. Sino-Tibetan, Japonic, and Austroasiatic languages see large gains (e.g., 40\% text for Sino-Tibetan), likely due to shared visual structures aiding transfer. Indo-European and Uralic languages show smaller gains ($<$5\%), suggesting their scripts were already well-modeled. Cyrillic gains are skewed toward tables, indicating structured content transfers more easily than text.

We provide more results on the effect of document type in appendix~\ref{appendix:doc-type-results}), markdown evaluation analysis in appendix~\ref{appendix:markdown-analysis}.

%% file: sec/6_conclusion.tex
\section{Conclusion}
We presented \textbf{DocAtlas}, a pipeline constructing multilingual OCR datasets through differential rendering, producing 360K training pages and a 5.8K-page benchmark across 82 languages and 9 tasks without learned models for core annotation. Evaluation of 14 models reveals persistent low-resource gaps and a language-invariant table ceiling (73\% TEDS), indicating spatial reasoning as the primary bottleneck. DPO with rendering-derived ground truth achieves stable cross-lingual transfer (+1.7\% accuracy, $<$3\% base degradation), outperforming stronger-model distillation, while QKV-only LoRA balances multilingual gains with capability preservation. Annotation quality, not model scale, bounds multilingual document understanding.

%% file: sec/7_limitations.tex
\section*{Limitations}

Our differential rendering pipeline requires access to native document source files (DOCX or structured markup), and therefore cannot annotate scanned or photographed documents lacking digital text layers. This is by design, model-free annotation relies on source-level structure, but limits applicability to born-digital corpora. Combining DocAtlas supervision with OCR-from-scratch methods for scanned documents is a natural extension.

\section*{Use of Language Models}
Large language models were used in a limited capacity to assist with minor editing and polishing of the manuscript, including improvements to clarity and grammar. All technical content, experimental design, results, and conclusions were produced, verified, and finalized by the authors.

%% file: sec/X_suppl.tex
\clearpage
\setcounter{page}{1}

\section{Data Generation Details}
\label{app:data_generation}

\subsection{Dataset Statistics and Composition}
\label{sec:app_stats}

The raw DocAtlas corpus contains 1{,}011{,}501 documents spanning 5.48M pages across 136 languages, sourced from two complementary pipelines: \textbf{Pipeline~A (Native DOCX)}, comprising 1{,}002{,}465 documents and 5.29M pages across 136 languages, and \textbf{Pipeline~B (Synthetic RTL)}, comprising 9{,}036 documents and 195K pages across 4 RTL languages (Arabic, Hebrew, Persian, Urdu). The data exhibits a long-tailed distribution: high-resource languages (English, Russian, Spanish) account for approximately 60\% of total pages, while medium- and low-resource scripts such as Hebrew, Thai, Burmese, and Khmer each contribute over 50{,}000 pages. After quality filtering and difficulty-aware sampling (\S\ref{sec:benchmark}), the final corpus comprises 360K training pages across 82 languages and a 5{,}862-page evaluation benchmark.

\begin{figure*}[t]
    \centering
    \includegraphics[width=\linewidth]{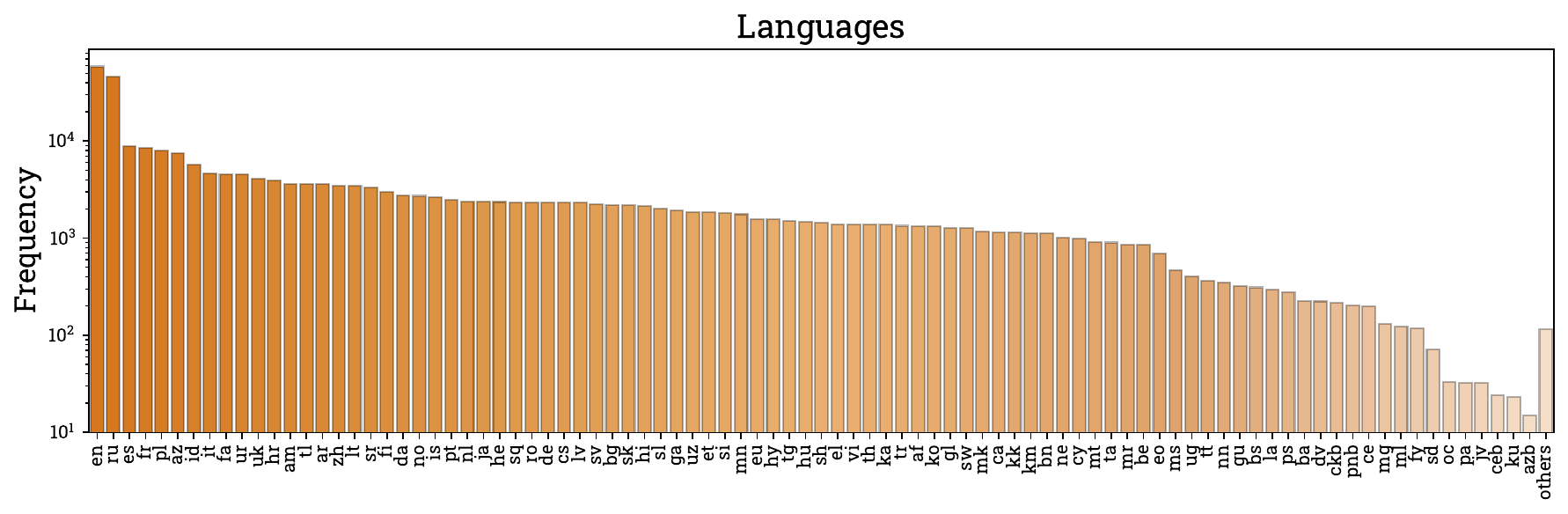}
    \caption{\textbf{Language frequency distribution in the DocAtlas corpus.} The dataset exhibits a long-tailed distribution across 80+ languages, with high-resource scripts (e.g., \texttt{en}, \texttt{ru}, \texttt{es}) dominating the head and low-resource languages (e.g., \texttt{ps}, \texttt{ckb}, \texttt{ku}, \texttt{azb}) forming a diverse tail.}
    \label{fig:lang-distribution}
\end{figure*}

\begin{figure}[t]
    \centering
    \includegraphics[width=\linewidth]{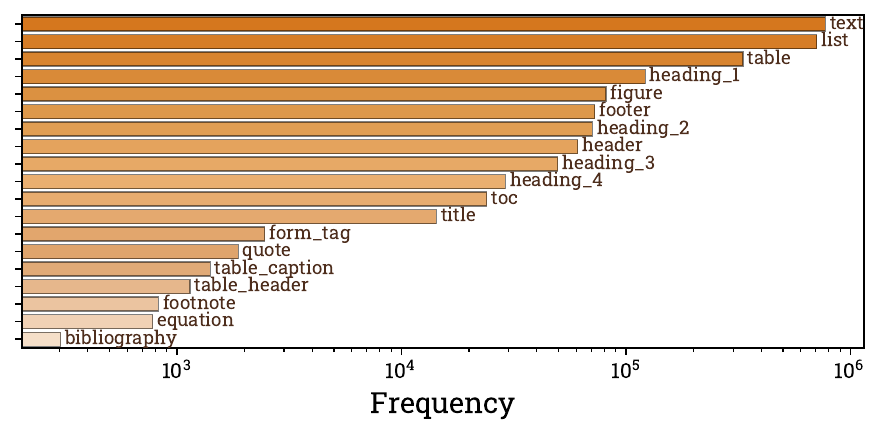}
    \caption{\textbf{Tag frequency distribution in DocAtlas.}}
    \label{fig:tags-distribution}
\end{figure}

Figure~\ref{fig:lang-distribution} illustrates the long-tailed language distribution in detail. English, Russian, and Spanish dominate the high-frequency tier, while medium- and low-resource scripts such as Hebrew, Thai, Burmese, and Khmer each contribute over 50{,}000 pages, ensuring meaningful representation across typological families. Figure~\ref{fig:tags-distribution} shows that DocAtlas spans over 30 structural element types. Dominant tags include \texttt{text}, \texttt{table}, and \texttt{heading\_1}, while less frequent yet critical tags such as \texttt{equation}, \texttt{form\_tag}, and \texttt{bibliography} provide supervision for rare but important document elements.
\subsection{Domain Diversity}
\label{sec:domain_diversity}

\begin{figure}[t]
    \centering
    \includegraphics[width=0.9\linewidth, trim=90 100 100 100, clip]{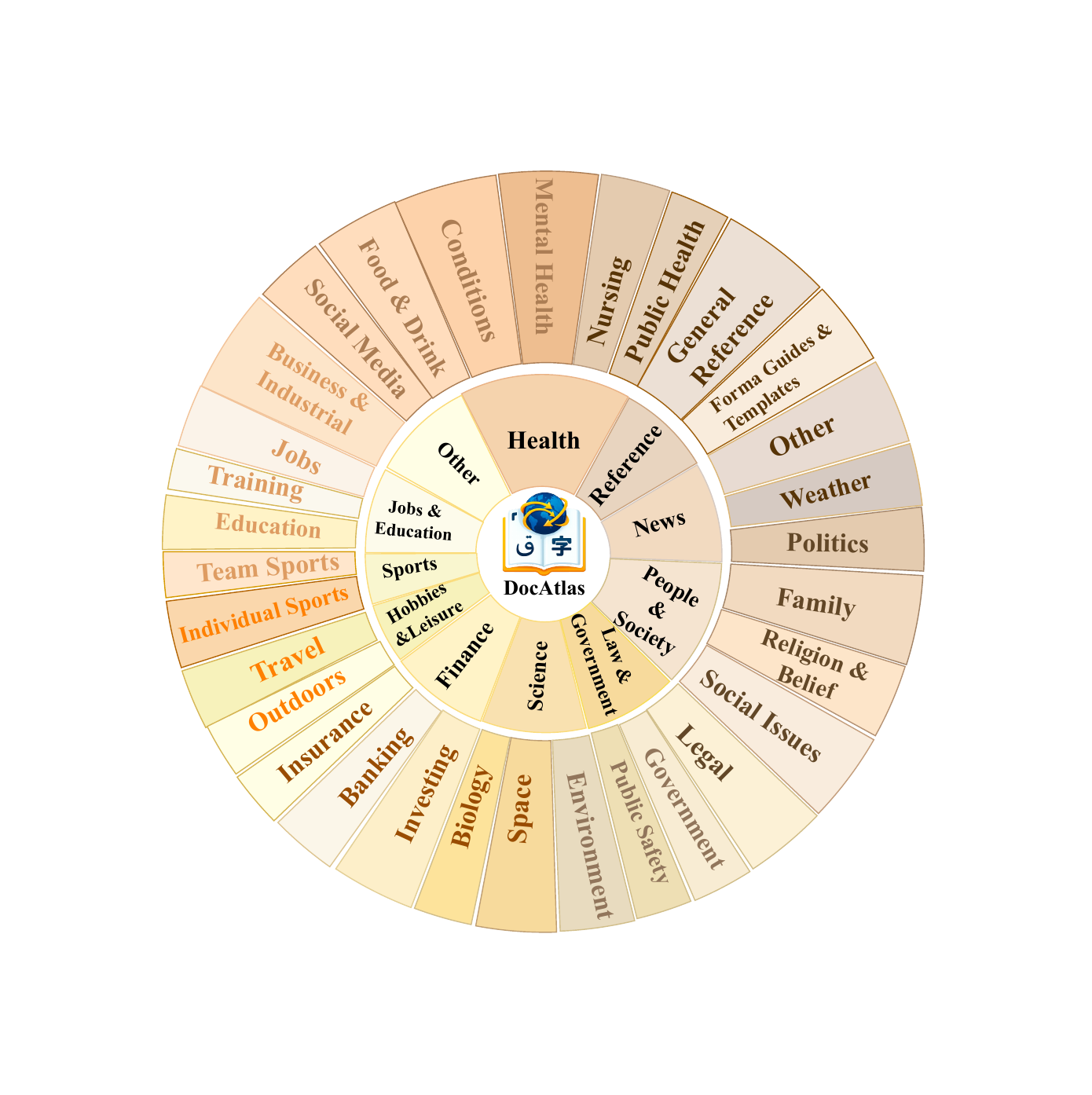}
    \caption{\textbf{Domain diversity across DocAtlas.} Each sector represents a major topic and its subdomains, showing balanced coverage across 25+ categories.}
    \label{fig:domain_diversity}
\end{figure}

To ensure broad generalization, the DocAtlas corpus spans over 25 primary categories and subcategories (Figure~\ref{fig:domain_diversity}), including \textit{Health}, \textit{Law \& Government}, \textit{Finance}, and \textit{Science}. Coverage is balanced across professional domains (e.g., \textit{Legal}, \textit{Investing}), academic domains (e.g., \textit{Biology}, \textit{Education}), and public-interest domains (e.g., \textit{Weather}, \textit{Religion \& Belief}), ensuring the corpus reflects both formal and informal document types encountered in practice.

\begin{table}[h]
    \centering
    \scriptsize
    \rowcolors{2}{white}{gray!10}
    \resizebox{\linewidth}{!}{
        \begin{tabular}{l l c}
            \toprule
            \textbf{Component} & \textbf{Method} & \textbf{Model-Free?} \\
            \midrule
            Text \& bounding boxes & Differential rendering & \cmark \\
            Tables \& structure & OpenXML tags & \cmark \\
            Captions \& reading order & XML cues + adjacency & \cmark \\
            \midrule
            Figure classification & Docling classifier & \xmark~(optional) \\
            Page attributes & Qwen3-VL & \xmark~(optional) \\
            \bottomrule
        \end{tabular}
    }
    \caption{\textbf{Model dependency breakdown.} Core structural annotations are fully model-free. Two optional enrichment steps use learned models but do not affect DocTag output.}
    \label{tab:model_dependency}
\end{table}

\subsection{Data License and Privacy Compliance}
\label{sec:license_privacy}

\paragraph{Licensing.}
All documents were sourced from publicly accessible web content and academic repositories under permissive licenses (CC-BY~4.0, CC0, public domain). Following established practices in web-scale dataset construction~\cite{nagel2023common,weber2023wordscape}, our crawler respects \texttt{robots.txt} directives and identifies itself via a documented user-agent with project contact information. Domains requiring opt-in permissions or explicitly prohibiting scraping were excluded, and no content was extracted from password-protected, paywalled, or authenticated sources.

\paragraph{Privacy Protection.}
We applied an automatic PII detection pipeline using Microsoft Presidio, configured with spaCy~3.5 and custom regex patterns for academic and professional documents. Table~\ref{tab:pii-summary} summarizes the detection categories and their precision. Documents containing three or more PII instances, or any government-issued identifier, were automatically excluded, resulting in 942{,}118 removals (5.15\% of 18.3M initially collected documents). Manual spot-checking of 1{,}000 retained documents confirmed a false-negative rate of 0.1\%.

\begin{table}[t]
\centering
\small

\rowcolors{2}{white}{gray!10}
\begin{tabular}{llc}
\toprule
\textbf{PII Category} & \textbf{Method} & \textbf{Precision} \\
\midrule
Personal names        & NER (spaCy)     & 95\% \\
Emails \& phone numbers & Regex         & 99\% \\
Government IDs        & Regex + rules   & 99\% \\
Addresses \& coordinates & NER + Regex  & 96\% \\
Financial identifiers & Regex           & 99\% \\
\bottomrule
\end{tabular}
\caption{\textbf{PII detection categories and precision.}}
\label{tab:pii-summary}
\end{table}

\subsection{Pipeline A: Native DOCX Generation Efficiency}
\label{sec:pipeline_a_details}

\paragraph{URL Extraction and Deduplication.}
We parsed three Common Crawl snapshots (CC-2023-14, CC-2023-06, CC-2021-43), extracting 11.4M raw URLs pointing to \texttt{.docx} and \texttt{.doc} files. Deduplication operated at two levels: (1)~within-snapshot URL canonicalization (lowercase, parameter sorting, trailing slash removal), which reduced each snapshot by 60--80\%, and (2)~cross-snapshot deduplication via a RocksDB key--value store with SHA-256 URL hashing, yielding 3.4M unique URLs.

\paragraph{Download, Safety Filtering, and Annotation.}
Table~\ref{tab:pipeline-a-funnel} traces the collection funnel from raw URLs to the final annotated corpus. Of 5.8M HTTP requests issued (with exponential-backoff retries), 42.1\% returned valid responses. Failures were dominated by dead links (38\%), timeouts (12\%), and redirects to non-document content (8\%). Downloaded files were further filtered for Content-Type mismatches (14.9\%) and malware indicators detected by OLETools (7.1\%), producing a clean set of 1.9M DOCX files. During OpenXML parsing and differential rendering, an additional 19.9\% of files were excluded due to corrupted ZIP archives (12.3\%), extremely short documents under 200 characters (4.8\%), and rendering engine crashes from unsupported features (2.8\%). The final output is 1{,}002{,}465 structurally annotated documents spanning 5.48M pages across 136 languages.

\begin{table}[t]
\centering
\small

\rowcolors{2}{white}{gray!10}
\resizebox{\linewidth}{!}{
\begin{tabular}{lrl}
\toprule
\textbf{Stage} & \textbf{Count} & \textbf{Primary Removals} \\
\midrule
Raw URLs extracted      & 11.4M & --- \\
After deduplication     & 3.4M  & 60--80\% per snapshot \\
Valid HTTP responses    & 2.44M & Dead links, timeouts \\
After safety filtering  & 1.9M  & Content-Type, malware \\
After parsing/rendering & 1.0M  & Corrupt ZIP, short docs \\
\bottomrule
\end{tabular}
}
\caption{\textbf{Pipeline~A collection funnel.} Each stage shows the surviving document count and the primary removal reasons.}
\label{tab:pipeline-a-funnel}
\end{table}

\paragraph{Throughput.}
All processing was executed on a single Apple M2-Pro (10-core CPU, 16GB RAM) without multiprocessing, due to the MS~Word singleton constraint. Table~\ref{tab:pipeline-a-throughput} reports per-stage timings. Sustained end-to-end throughput exceeds 100K annotated pages per day, completing 100K samples per snapshot in under 72 hours without GPU acceleration or distributed computing.

\begin{table}[h]
\centering
\scriptsize
\rowcolors{2}{white}{gray!10}
\resizebox{\linewidth}{!}{
\begin{tabular}{@{}lrl@{}}
\toprule
\textbf{Operation} & \textbf{Time} & \textbf{Throughput} \\
\midrule
URL extraction          & 0.02s & 4{,}320K pg/day \\
Doc.\ download          & 1.20s & 72K pg/day \\
OpenXML parsing         & 0.15s & 576K pg/day \\
Diff.\ rendering        & 0.45s & 192K pg/day \\
Text extract.\ \& align & 0.12s & 720K pg/day \\
\midrule
\textbf{Total (E2E)}   & \textbf{0.74s} & \textbf{117K pg/day} \\
\bottomrule
\end{tabular}
}
\caption{\textbf{Per-page processing time for Pipeline~A.}}
\label{tab:pipeline-a-throughput}
\end{table}

\begin{figure}[tp]
    \centering
    \includegraphics[width=\linewidth]{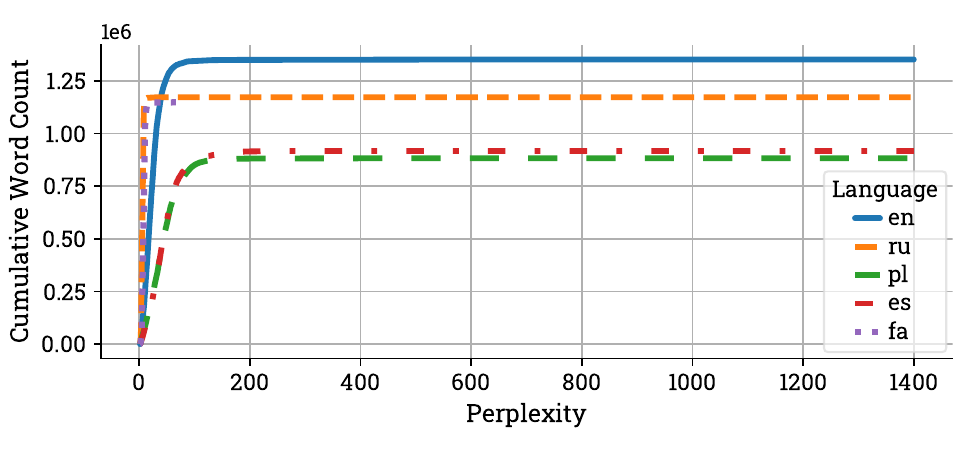}
    \caption{\textbf{Perplexity-based filtering across five languages.}}
    \label{fig:perplexity_filtering}
\end{figure}

\paragraph{Quality Filtering and Perplexity Analysis.}\label{appendix:quality-filtering}
To maintain high multilingual quality, we apply a two-stage filtering process. First, we predict document language using fastText~\cite{fasttext} and compute perplexity via language-specific 5-gram Kneser–Ney models~\cite{ccnet}. As shown in Figure~\ref{fig:perplexity_filtering}, 
thresholding at $\tau = 120$ retains over 94\% of high-quality data while filtering out 38\% of low-confidence pages. Second, we compute an \textit{annotation reliability score}, defined as the proportion of characters successfully tagged via XML styles or table markers rather than heuristics. Pages with reliability below 0.6 or with anomalous visual signals (excessive blank space, corrupted rendering) are excluded, resulting in roughly 15\% removal of low-confidence pages following~\cite{weber2023wordscape}.


\subsection{Pipeline B: Synthetic RTL Data Generation}
\label{sec:pipeline_b_details}

\paragraph{Source Documents and Output.}
We collected structured documents (EPUB, HTML, XML) from open digital libraries and academic repositories for four RTL languages. Table~\ref{tab:pipeline-b-summary} summarizes the input sources, template counts, and filtered output per language. The average document length is 3.5 pages.

\begin{table}[h]
\centering
\small

\rowcolors{2}{white}{gray!10}
\resizebox{\linewidth}{!}{
\begin{tabular}{lrrrc}
\toprule
\textbf{Lang.} & \textbf{Source Docs} & \textbf{Templates} & \textbf{Output Pages} & \textbf{\%} \\
\midrule
Arabic  & 4{,}636 & 64 & 123K  & 63.1 \\
Hebrew  & 2{,}937 & 60 & 39K   & 20.0 \\
Persian & 1{,}378 & 36 & 26.3K & 13.5 \\
Urdu    & 85      & 45 & 6.7K  & 3.4  \\
\midrule
\textbf{Total} & 9{,}036 & 205 & 195K & 100 \\
\bottomrule
\end{tabular}
}
\caption{\textbf{Pipeline~B: RTL data summary by language.}}
\label{tab:pipeline-b-summary}
\end{table}

\paragraph{Template System.}
Each language-specific \LaTeX{} template defines page format (portrait/landscape), column count (1--3), font family and size (9--14pt), colors, margins, header/footer styles, and bidirectional text handling primitives. Font selections reflect typographic conventions of each script: Amiri, Scheherazade, and Traditional Arabic for Arabic; David, Narkisim, and Frank Ruehl for Hebrew; Nazanin, Lotus, and Iranian Sans for Persian; and Nastaliq and Naskh for Urdu. Persian templates additionally support mixed LTR/RTL layouts for scientific content.

\paragraph{Rendering and Quality Control.}
The synthesis engine, built on LuaTeX with custom positional logging commands, operates in three compilation passes: (1)~initial layout with placeholder positions, (2)~position logging that writes exact element coordinates to \texttt{.pos} files, and (3)~final rendering with validated positions. Sustained throughput is 183~pages/minute (10{,}980~pages/hour) on a single CPU core.

Quality filtering removes pages with positional inconsistencies exceeding 2pt coordinate drift between passes (15.2\% of raw output), template misalignments such as overlapping elements or text overflow (8.9\%), and font rendering failures including missing glyphs or incorrect shaping (2.1\%). The filtered output of 195K pages demonstrates the scalability of the pipeline for producing bidirectional, script-accurate OCR supervision.

\paragraph{Input Parsing.}
Structured inputs are parsed into a standardized Docling JSON schema using BeautifulSoup\footnote{https://beautiful-soup-4.readthedocs.io/en/latest/} and ebooklib\footnote{https://docs.sourcefabric.org/projects/ebooklib/en/latest/}. Each content element, headers, paragraphs, tables, and figures, is tagged and assigned provisional bounding boxes, which are later refined during \LaTeX{} compilation.

\paragraph{Template Engine.}
The template engine governs typography (font family, size, color), layout (single or multi-column, margins, spacing), and positioning. We implement 205 LuaTeX-based templates covering Arabic, Hebrew, Urdu, and Persian, each supporting bidirectional text control. For mixed-language segments, Latin text is wrapped in explicit bidirectional control markers (\texttt{\textbf{\textcolor{mycmd}{\textbackslash textdir TLT}}} \ldots\ \texttt{\textbf{\textcolor{mycmd}{\textbackslash textdir TRT}}}) to preserve visual ordering. The pipeline also handles structured tables with \textbf{\textcolor{mycmd}{\texttt{colspan}}} and \textbf{\textcolor{mycmd}{\texttt{rowspan}}} attributes, and programmatically generates charts (bar, pie, or stacked) from tabular data using Matplotlib\footnote{https://matplotlib.org/} or Seaborn\footnote{https://seaborn.pydata.org/}.

\paragraph{Coordinate Recovery.}
During compilation, custom \LaTeX{} commands log positional metadata for each element, allowing exact bounding-box recovery. Coordinates in scaled points (sp) are successively transformed across coordinate systems: \textit{LaTeX (bottom-left, sp)} $\rightarrow$ \textit{PDF (bottom-left, pt)} $\rightarrow$ \textit{Image (top-left, px)}. Three compilation passes, initial layout, position logging, and final rendering, guarantee positional stability. The resulting output pairs a rendered PDF with Docling~\cite{docling} JSON containing element-level bounding boxes, text content, structural labels, and annotation confidence scores.

\subsection{Dataset Statistics and Quality Control}
We sourced 1.9M documents spanning 5.48M pages across 136 languages from Commoncrawl, with all data sourced from publicly accessible web content under permissive licenses. Automated PII detection removed 5.15\% of documents containing sensitive content. Our native pipeline (Pipeline A) processes 100k+ pages/day, while the synthetic RTL pipeline (Pipeline B) generated 195k pages at 183 pages/minute. 
Additionally, our differential rendering pipeline achieves near-perfect annotation fidelity in most cases. However, three document classes require targeted filtration: (i)~\textit{Scanned PDFs} rendered as monolithic images without structural markup are detected via metadata analysis and excluded; (ii)~\textit{Rendering drift} from missing fonts ($<$0.3\% incidence) causes subpixel geometry shifts, mitigated through tolerance-aware contour matching; (iii)~\textit{Malformed OpenXML} documents with corrupted namespaces are repaired via schema validation and XML normalization, with unrecoverable cases filtered post-validation. These quality controls ensure annotation precision across the retained corpus. After quality filtering and difficulty-aware sampling, these source documents yielded a curated corpus of 360k training pages across 82 languages. The resulting corpus exhibits diverse language and component distributions, with 31 structural element types spanning from high-frequency tags (\textbf{\textcolor{mycmd}{\texttt{text}}}, \textbf{\textcolor{mycmd}{\texttt{table}}}) to rare but critical elements (\textbf{\textcolor{mycmd}{\texttt{equation}}}, \textbf{\textcolor{mycmd}{\texttt{bibliography}}}). Lastly, to ensure broad generalization, the corpus spans 25+ domains including Health, Law \& Government, Finance, and Science, with balanced coverage across professional, academic, and public interest categories.

\subsection{Benchmark Construction Details}\label{appendix:benchmark}

\paragraph{Difficulty-Stratified Sampling.}
Following the clustering protocol of~\cite{omnidocbench}, each page is embedded with ResNet-50~\cite{resnet} features and clustered via FAISS~\cite{faiss} to promote visual variety. Within each cluster, a difficulty score combines per-component weights (e.g., table, formula, chart, density, font variety, image ratio), and sampling is drawn from a normal distribution to approximate equal easy, medium, and hard splits. This yields up to 100 pages per language across 82 languages (5,575 samples). To enrich the benchmark, we manually curate 201 additional PDFs containing challenging formulas, adding 144 samples.

\paragraph{Chart Data Generation.}
To enrich our benchmark with charts, we develop a pipeline inspired by~\cite{kitabbench}. Topics are generated using an expert VLM (Qwen3-8B-VL~\cite{qwen3}), rendered to multiple chart types via Matplotlib or Plotly\footnote{https://plotly.com/python/}, and filtered using GPT-4o~\cite{gpt4o} before expert verification by three domain specialists who cross-verified structural integrity, \LaTeX{} formula alignment, and RTL reading order, achieving inter-annotator agreement of 94.2\% ($\kappa$=0.89).

\paragraph{Page-Level Attributes.}
We further annotate pages with fine-grained attributes such as column layout, watermark presence, and background color, predicted via Qwen3-8B-VL and confirmed by human inspection. These attributes enable controlled evaluations under specific visual or layout conditions.

\subsection{Annotation Failure Analysis}
\label{sec:failure_analysis}

Although our differential rendering pipeline achieves near-perfect fidelity (98.9\% of documents maintain ${>}95\%$ annotation accuracy), three document classes consistently degrade annotation quality and require targeted intervention: scanned or rasterized PDFs (8.2\% of downloads, excluded), rendering drift from font substitution (${<}0.3\%$, largely mitigated), and malformed OpenXML markup (repaired where possible).

\paragraph{Scanned or Rasterized PDFs.}
Documents consisting of page-sized raster images without text layers yield only a single \texttt{<image>} tag and provide no structured supervision. We detect them via a two-stage process: metadata analysis (checking for full-page \texttt{/Type /Image} entries or absence of text operators) followed by image entropy thresholding (scanned content typically exceeds 6.5 vs.\ below 4.5 for rendered text). Approximately 8.2\% of downloaded PDFs are flagged and excluded; these are logged separately for potential use in OCR-from-scratch benchmarks.

\paragraph{Rendering Drift.}
Missing or non-embedded fonts trigger substitution fallbacks that introduce subpixel horizontal shifts, line-wrap changes, and vertical spacing adjustments between the colorized and uncolorized renderings. Although rare (${<}0.3\%$ overall), these cases are concentrated in mathematical papers and financial reports (10--15\% of those categories). Our mitigation pipeline applies layout fingerprinting to detect structural mismatches between renderings, tolerance-aware contour matching with relaxed IoU and pixel-offset thresholds, and subpixel registration via phase correlation before differencing. Pages with residual misalignment exceeding 5\% of components are conservatively excluded. This recovers 92\% of affected documents; the remaining 8\% (0.024\% of the total corpus) are filtered.

\paragraph{Malformed OpenXML.}
Documents with unclosed or misordered tags, duplicated element IDs, or corrupted namespace declarations often pass basic ZIP integrity checks but fail during deterministic element-to-geometry mapping, producing symptoms such as misclassified tags, silent annotation loss, or incorrect nesting. We intercept these via ECMA-376 strict schema validation, apply automatic XML repair (tag balancing, ID deduplication, namespace normalization), and verify consistency by comparing expected versus recovered element counts after rendering. Documents whose annotation reliability score, the proportion of characters tagged via native XML signals rather than heuristics, falls below 0.6 are excluded. This process successfully repairs 73\% of malformed documents; the remaining 27\% (0.8\% of the total corpus) are filtered. The repair step adds 0.08s average overhead per document.

\paragraph{Quality Validation.}
To verify annotation accuracy across all filtering stages, we manually spot-checked 500 random samples per language (70{,}000 total pages across 140 languages). Human annotators marked bounding boxes and labels as correct or incorrect, computing per-page accuracy as:
\[
\text{Accuracy} = \frac{\text{Correct annotations}}{\text{Total annotations}} \times 100\%
\]
Results show that 98.9\% of retained documents achieve ${>}95\%$ annotation accuracy, with a mean of 97.8\% ($\sigma = 2.1\%$) across the corpus. The 1.1\% of documents falling below 95\% are retained with warning flags.

\section{Experiments}

\subsection{Metric Definitions}\label{sec:metrics}

We evaluate model outputs using five complementary metrics spanning text, table, formula, chart, and full-page performance.

\paragraph{Normalized Edit Distance (NED).}
The NED metric measures the similarity between a predicted string~$p$ and a ground-truth string~$g$:
\begin{equation}
\text{NED}(p, g) = 1 - \frac{D_{\text{Lev}}(p, g)}{\max(|p|, |g|)}
\end{equation}
where $D_{\text{Lev}}(p, g)$ is the Levenshtein edit distance. NED ranges from~0 (completely dissimilar) to~1 (identical). We report $1 - \text{NED}$ as \textit{TextEdit} in Table~3, so lower is better.

\paragraph{Tree-Edit-Distance-based Similarity (TEDS).}
For table evaluation, predicted and ground-truth outputs are converted to HTML and parsed as DOM trees:
\begin{equation}
\text{TEDS}(T_p, T_g) = 1 - \frac{D_{\text{tree}}(T_p, T_g)}{\max(|T_p|, |T_g|)}
\end{equation}
where $D_{\text{tree}}$ denotes the tree edit distance between DOM trees $T_p$ and~$T_g$. TEDS penalizes both structural errors (missing rows, merged cells) and textual discrepancies within cells.

\paragraph{Character Detection Matching (CDM).}
For formula evaluation, CDM measures spatial alignment between detected and ground-truth characters:
\begin{align}
\text{Precision} &= \frac{|\mathcal{M}|}{|\mathcal{P}|}, \quad
\text{Recall} = \frac{|\mathcal{M}|}{|\mathcal{G}|} \\
\text{CDM} &= 100 \times \frac{2 \times \text{Precision} \times \text{Recall}}{\text{Precision} + \text{Recall}}
\end{align}
where $\mathcal{P}$ and $\mathcal{G}$ are sets of predicted and ground-truth characters, and $\mathcal{M}$ denotes matched character pairs under a spatial or syntactic tolerance.

\paragraph{Chart Score.}
Since charts are first converted into structured HTML tables, we evaluate chart extraction accuracy using the same TEDS metric applied to the resulting table representations:
\begin{equation}
\text{ChartScore}(c_p, c_g) = \text{TEDS}\!\bigl(\texttt{html}(c_p),\; \texttt{html}(c_g)\bigr)
\end{equation}
where $\texttt{html}(\cdot)$ denotes the chart-to-HTML-table conversion. This formulation captures both the structural layout (rows, columns, headers) and the textual content of the underlying data.

\subsection{Setup}\label{appendix:setup}

All training experiments use 4$\times$A100 GPUs (40GB each), effective batch size 32, 3 epochs with cosine-scheduled learning rate of $2 \times 10^{-5}$, AdamW optimizer, LoRA rank 16 ($\alpha$=32), and DPO $\beta$=0.1. Pipeline A throughput benchmarks were measured on a single Apple M2-Pro (10-core CPU, 16GB RAM), sustaining 100K+ annotated pages/day without GPU acceleration. To disentangle the contribution of our annotation pipeline from the DPO algorithm itself, we additionally compare against a distillation baseline where GPT-4o outputs serve as positive examples (\S\ref{sec:results}).

\begin{figure}[h]
    \centering
    \includegraphics[width=\linewidth]{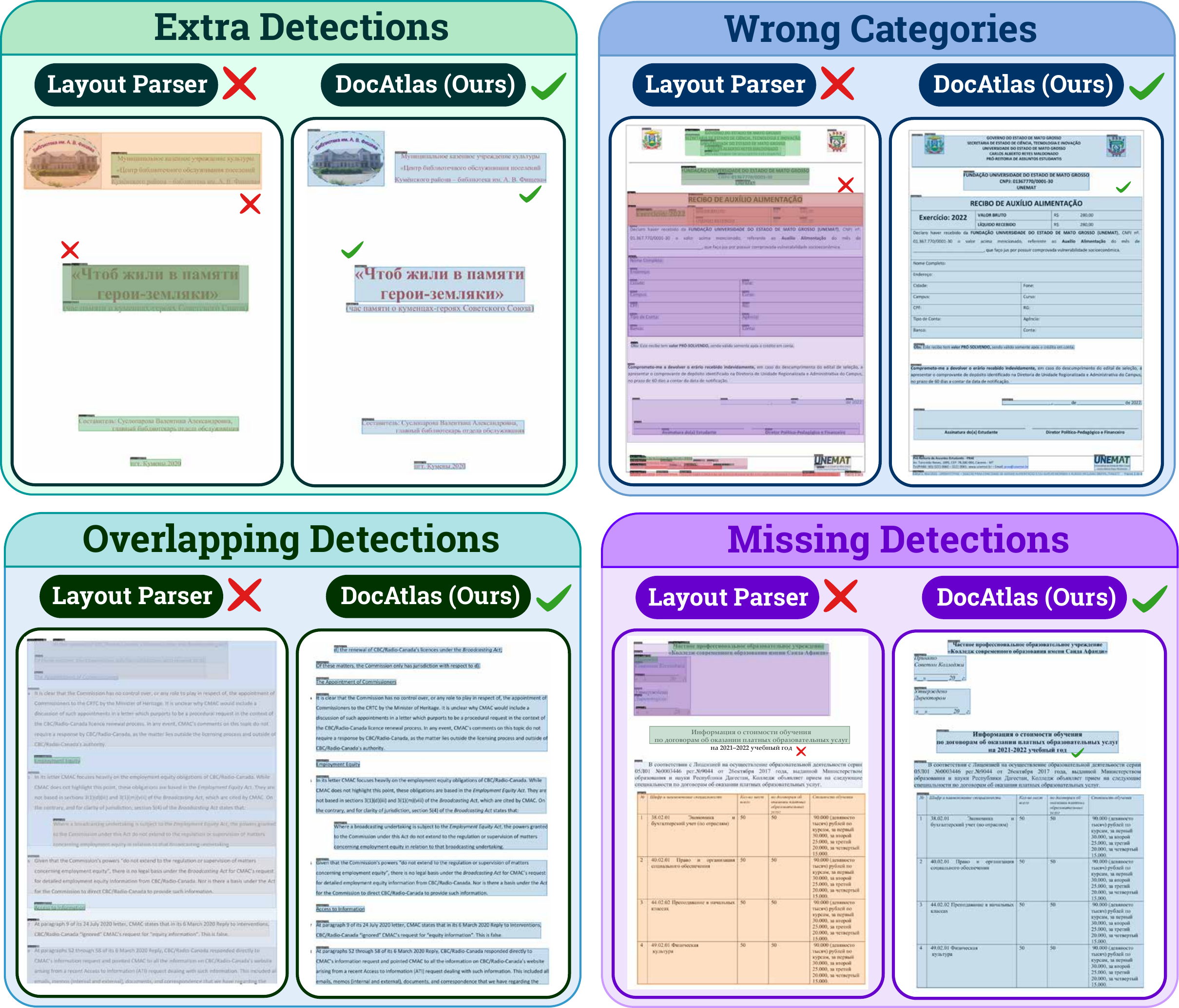}
    \caption{\textbf{Qualitative comparison of layout parser failure cases.} Common error types (extra, overlapping, missing detections; wrong categories) are shown for Layout Parser (left) and our DocAtlas system (right). DocAtlas consistently produces more accurate and cleaner segmentation.}
    \label{fig:layout_failures}
\end{figure}

\subsection{Layout Robustness}\label{appendix:layout-robust}  Figure~\ref{fig:layout_failures} provides a qualitative comparison between traditional layout parsers~\cite{docling,layoutlmv3,shen2021layoutparser} and the proposed DocAtlas annotation pipeline. Layout parsers often fail under visually complex conditions such as overlapping elements, nonstandard styles, or RTL scripts, resulting in fragmented bounding boxes and incorrect component classification. In contrast, DocAtlas maintains structural integrity even in pages with dense tables, multilingual content, or rotated elements, owing to its differential rendering and OpenXML-based annotation.

\subsection{Chart Extraction Evaluation Setup}
\label{appendix:chart_setup}

To assess cross-lingual robustness in chart understanding, we evaluated OCR and vision-language models with native chart parsing capabilities. The evaluated systems include DeepseekOCR~\cite{deepseekocr2025}, SmolDocling~\cite{smoldocling}, NanosetsOCR2~\cite{nanonetsocr2}, Gemini-2.5-Flash~\cite{gemini25}, GPT-4o, and GPT-4o-mini~\cite{gpt4o}. We selected 15 representative languages: Arabic, Chinese, Croatian, Dutch, English, French, Hindi, Italian, Polish, Russian, Serbian, Spanish, Thai, Ukrainian, and Vietnamese. These languages cover diverse writing systems including Latin, Cyrillic, Arabic, Devanagari, and logographic scripts, representing both high-resource and regionally significant languages for real-world multilingual scenarios.

\section{Results}

\begin{figure}[t]
    \centering
    \includegraphics[width=\linewidth]{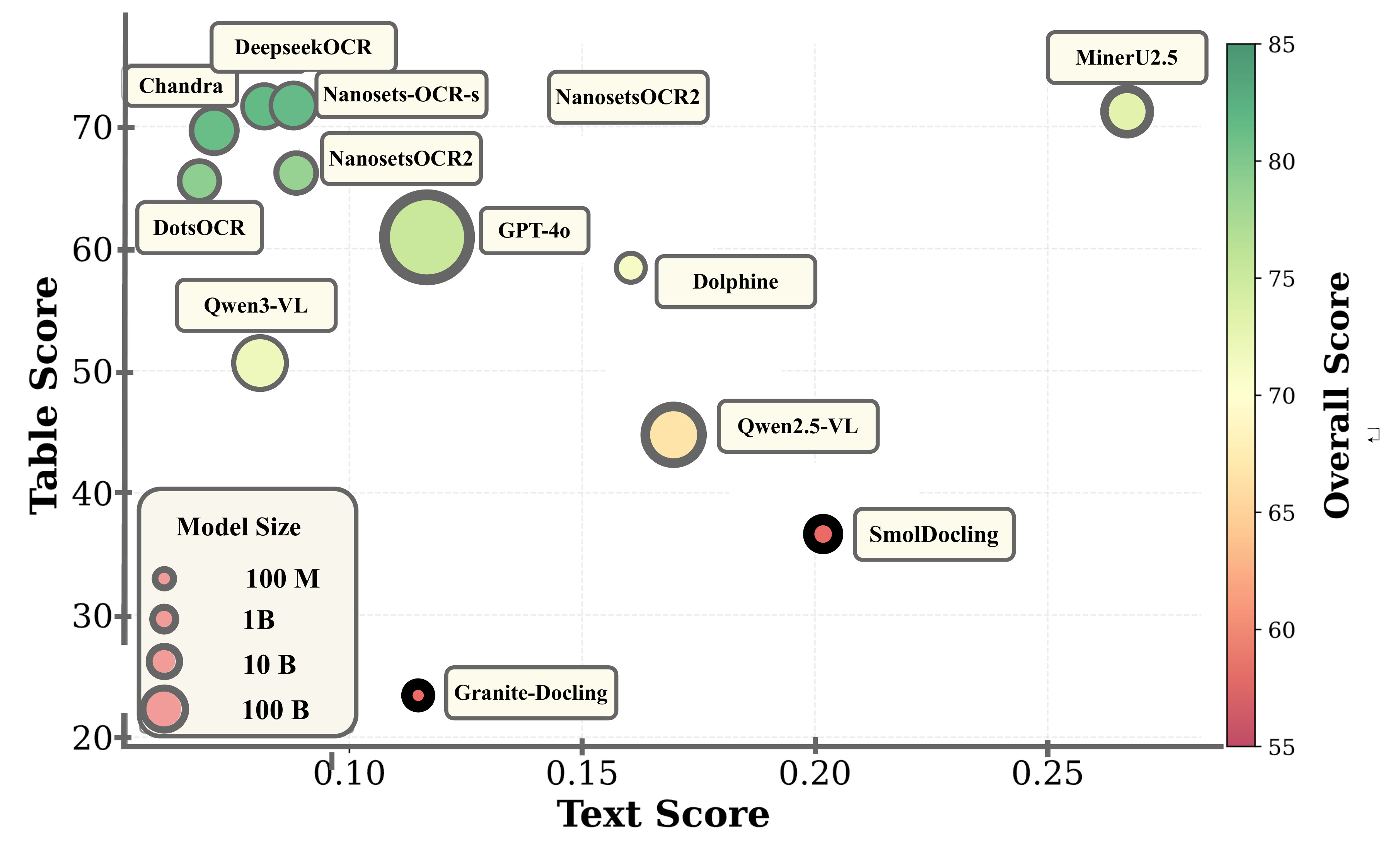}
    \caption{\textbf{Scores vs.\ model scale.} Each point represents a model; marker size encodes parameter count. Larger models do not uniformly dominate: several compact expert systems ($\leq$3B) match or exceed general-purpose VLMs on both text and table scores.}
    \label{fig:score-vs-scale}
\end{figure}

\subsection{Scores vs.\ Model Scale}\label{appendix:scores-vs-scale}
Figure~\ref{fig:score-vs-scale} visualizes the relationship between model scale and overall performance, showing that compact expert systems can rival much larger general-purpose VLMs.

\begin{figure}[h]
    \centering
    \includegraphics[width=0.8\linewidth]{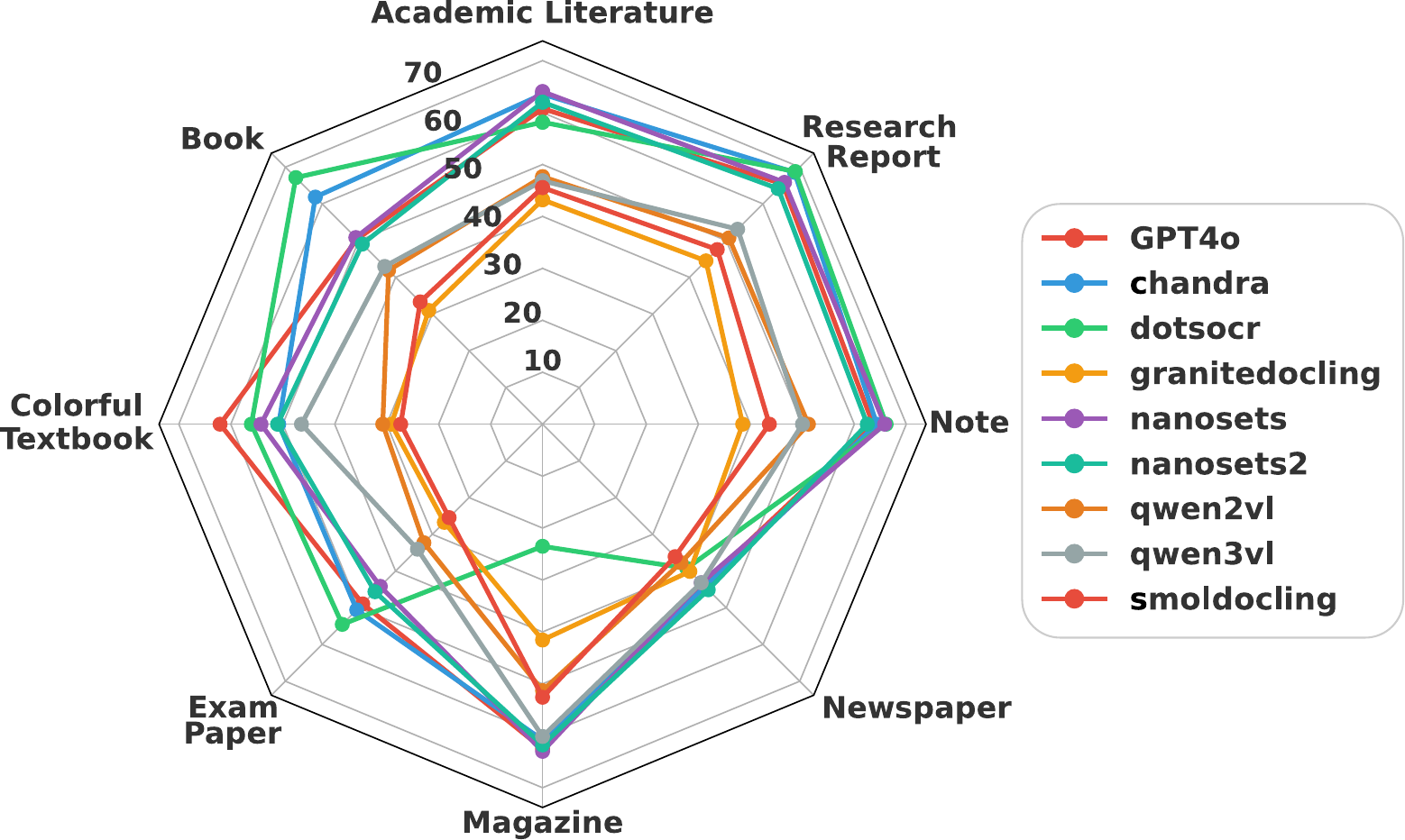}
    \caption{\textbf{Document type performance.} }
    \label{fig:results-per-documents}
\end{figure}

\subsection{Document Type Evaluation}\label{appendix:doc-type-results} (Figure~\ref{fig:results-per-documents}) confirms training data composition directly shapes capabilities: models excel on academic literature and research reports (65-70\% accuracy), domains heavily represented in datasets like DocBank~\cite{li2020docbank}, yet collapse on newspapers and magazines (30-45\%), where dense multi-column layouts and varied typography break learned priors. This performance asymmetry indicates that benchmark diversity, not just scale, determines real-world generalization.

\subsection{Markdown Error Analysis}\label{appendix:markdown-analysis}
Systematic evaluation of 11 OCR models across 5,345 documents reveals 88,036 errors across 12 categories, with several dominant patterns.Spacing errors are most frequent (15.7\%), primarily involving extraneous line breaks that fragment paragraphs. Formatting errors (14.6\%) manifest as incorrect bold/italic tags and inconsistent dash characters (U+2212 vs U+2013). Character encoding errors (13.2\%) involve Unicode normalization issues, particularly with ellipsis characters. Content omission (13.2\%) affects hyphenated words and list separators. Structural challenges include table structure errors (8.3\%) with incorrect \texttt{<thead>} insertion and list formatting errors (8.2\%) with inappropriate \texttt{<br>} tags. Heading errors (7.1\%) show systematic level misclassification, particularly H6 to H1/H2 conversions. Lower-frequency issues include content addition (7.1\%), link errors (0.2\%), image reference errors (0.7\%), code block errors (0.6\%), and math formula errors (0.1\%).

%% file: custom.bib
@String(ICLR = {Int. Conf. Learn. Represent.})

@String(AAAI = {AAAI})

@String(ICLR  = {ICLR})

@article{li2025monkeyocr,
  title={MonkeyOCR: Document Parsing with a Structure-Recognition-Relation Triplet Paradigm},
  author={Li, Zhang and Liu, Yuliang and Liu, Qiang and Ma, Zhiyin and Zhang, Ziyang and Zhang, Shuo and Guo, Zidun and Zhang, Jiarui and Wang, Xinyu and Bai, Xiang},
  journal={arXiv preprint arXiv:2506.05218},
  year={2025}
}

@article{deepseekocr2025,
  title={Deepseek-ocr: Contexts optical compression},
  author={Wei, Haoran and Sun, Yaofeng and Li, Yukun},
  journal={arXiv preprint arXiv:2510.18234},
  year={2025}
}

@inproceedings{blecher2023nougat,
  title={Nougat: Neural optical understanding for academic documents},
  author={Blecher, Lukas and Cucurull Preixens, Guillem and Scialom, Thomas and Stojnic, Robert},
  booktitle={International Conference on Learning Representations},
  volume={2024},
  pages={37646--37663},
  year={2024}
}

@article{auer2024docling,
  title={Docling technical report},
  author={Auer, Christoph and Lysak, Maksym and Nassar, Ahmed and Dolfi, Michele and Livathinos, Nikolaos and Vagenas, Panos and Ramis, Cesar Berrospi and Omenetti, Matteo and Lindlbauer, Fabian and Dinkla, Kasper and others},
  journal={arXiv preprint arXiv:2408.09869},
  year={2024}
}

@misc{granitedocling,
  title={Granite Docling: A 258M-Parameter Multimodal VLM for Document Understanding},
  author={{IBM Granite Team}},
  howpublished={\url{https://huggingface.co/ibm-granite/granite-docling-258M}},
  year={2025}
}

@inproceedings{smoldocling,
  title={SmolDocling: An ultra-compact vision-language model for end-to-end multi-modal document conversion},
  author={Nassar, Ahmed and Omenetti, Matteo and Lysak, Maksym and Livathinos, Nikolaos and Auer, Christoph and Morin, Lucas and de Lima, Rafael Teixeira and Kim, Yusik and Gurbuz, A Said and Dolfi, Michele and others},
  booktitle={Proceedings of the IEEE/CVF International Conference on Computer Vision},
  pages={21972--21983},
  year={2025}
}

@inproceedings{mineruv25,
  title={Mineru2. 5: A decoupled vision-language model for efficient high-resolution document parsing},
  author={Niu, Junbo and Liu, Zheng and Gu, Zhuangcheng and Wang, Bin and Ouyang, Linke and Zhao, Zhiyuan and Chu, Tao and He, Tianyao and Wu, Fan and Zhang, Qintong and others},
  booktitle={The 64th Annual Meeting of the Association for Computational Linguistics--Industry Track},
  year={2025}
}

@misc{dotsocr2025,
  title={dots.ocr: Multilingual Document Layout Parsing in a Single Vision-Language Model},
  author={{Xiaohongshu Hi Lab}},
  howpublished={\url{https://github.com/rednote-hilab/dots.ocr}},
  year={2025}
}

@inproceedings{feng2025dolphin,
  title={Dolphin: Document Image Parsing via Heterogeneous Anchor Prompting},
  author={Feng, Hao and Wei, Shu and Fei, Xiang and Shi, Wei and Han, Yingdong and Liao, Lei and Lu, Jinghui and Wu, Binghong and Liu, Qi and Lin, Chunhui and Tang, Jingqun and Liu, Hao and Huang, Can},
  booktitle={Proceedings of the 65th Annual Meeting of the Association for Computational Linguistics (ACL)},
  year={2025},
  note={arXiv:2505.14059}
}

@misc{nanonetsocr2,
  title={Nanonets-OCR2: A model for transforming documents into structured markdown with intelligent content recognition and semantic tagging},
  author={Souvik Mandal and Ashish Talewar and Siddhant Thakuria and Paras Ahuja and Prathamesh Juvatkar},
  year={2025},
}

@misc{paddleocrvl,
  title={PaddleOCR-VL: Boosting Multilingual Document Parsing via a 0.9B Ultra-Compact Vision-Language Model},
  author={Cui, C. and others},
  howpublished={ERNIE Technical Report},
  year={2025},
  note={Available at \url{https://ernie.baidu.com/}}
}

@inproceedings{jaume2019funsd,
  title={FUNSD: A Dataset for Form Understanding in Noisy Scanned Documents},
  author={Jaume, Guillaume and Ekenel, Hazim Kemal and Thiran, Jean-Philippe},
  booktitle={2019 International Conference on Document Analysis and Recognition Workshops (ICDARW)},
  volume={2},
  pages={1--6},
  year={2019},
  organization={IEEE},
  doi={10.1109/ICDARW.2019.10029},
  note={arXiv:1905.13538}
}

@inproceedings{xu2022xfund,
  title={XFUND: A Benchmark Dataset for Multilingual Visually Rich Form Understanding},
  author={Xu, Yiheng and Lv, Tengchao and Cui, Lei and Wang, Guoxin and Lu, Yijuan and Florencio, Dinei and Zhang, Cha and Wei, Furu},
  booktitle={Findings of the Association for Computational Linguistics: ACL 2022},
  pages={3214--3224},
  year={2022},
  doi={10.18653/v1/2022.findings-acl.253}
}

@inproceedings{zhong2019publaynet,
  title={PubLayNet: Largest Dataset Ever for Document Layout Analysis},
  author={Zhong, Xu and Tang, Jianbin and Yepes, Antonio Jimeno},
  booktitle={2019 International Conference on Document Analysis and Recognition (ICDAR)},
  pages={1015--1022},
  year={2019},
  organization={IEEE},
  doi={10.1109/ICDAR.2019.00166},
  note={arXiv:1908.07836}
}

@article{pfitzmann2022doclaynet,
  title={DocLayNet: A Large Human-Annotated Dataset for Document-Layout Analysis},
  author={Pfitzmann, Birgit and Auer, Christoph and Dolfi, Michele and Nassar, Ahmed S. and Staar, Peter},
  journal={arXiv preprint arXiv:2206.01062},
  year={2022}
}

@article{liu2024ocrbench,
  title={OCRBench: On the Hidden Mystery of OCR in Large Multimodal Models},
  author={Liu, Yuliang and others},
  journal={arXiv preprint arXiv:2305.07895},
  year={2024}
}

@inproceedings{shen2021layoutparser,
  title={LayoutParser: A Unified Toolkit for Deep Learning Based Document Image Analysis},
  author={Shen, Zejiang and Zhang, Ruochen and Dell, Melissa and Lee, Benjamin Charles Germain and Carlson, Jacob and Li, Weining},
  booktitle={Document Analysis and Recognition--ICDAR 2021: 16th International Conference, Lausanne, Switzerland, September 5--10, 2021, Proceedings, Part I 16},
  pages={131--146},
  year={2021},
  organization={Springer},
  doi={10.1007/978-3-030-86549-8_9},
  note={arXiv:2103.15348}
}

@article{xu2021layoutxlm,
  title={LayoutXLM: Multimodal Pre-training for Multilingual Visually-Rich Document Understanding},
  author={Xu, Yiheng and Lv, Tengchao and Cui, Lei and Wang, Guoxin and Lu, Yijuan and Florencio, Dinei and Zhang, Cha and Wei, Furu},
  journal={arXiv preprint arXiv:2104.08836},
  year={2021}
}

@inproceedings{yim2021synthtiger,
  title={SynthTIGER: Synthetic Text Image GEneratoR Towards Better Text Recognition Models},
  author={Yim, Moonbin and Kim, Yoonsik and Cho, Han-Cheol and Park, Sungrae},
  booktitle={Document Analysis and Recognition--ICDAR 2021: 16th International Conference, Lausanne, Switzerland, September 5--10, 2021, Proceedings, Part III 16},
  pages={109--124},
  year={2021},
  organization={Springer},
  doi={10.1007/978-3-030-86337-1_8},
  note={arXiv:2107.09313}
}

@article{journet2017doccreator,
  title={DocCreator: A New Software for Creating Synthetic Ground-Truthed Document Images},
  author={Journet, Nicholas and Visani, Muriel and Mansencal, Boris and Van-Cuong, Kieu and Billy, Antoine},
  journal={Journal of Imaging},
  volume={3},
  number={4},
  pages={62},
  year={2017},
  publisher={MDPI},
  doi={10.3390/jimaging3040062}
}

@inproceedings{omnidocbench,
  title={Omnidocbench: Benchmarking diverse pdf document parsing with comprehensive annotations},
  author={Ouyang, Linke and Qu, Yuan and Zhou, Hongbin and Zhu, Jiawei and Zhang, Rui and Lin, Qunshu and Wang, Bin and Zhao, Zhiyuan and Jiang, Man and Zhao, Xiaomeng and others},
  booktitle={Proceedings of the Computer Vision and Pattern Recognition Conference},
  pages={24838--24848},
  year={2025}
}

@article{qwen25,
  title={Qwen2. 5-vl technical report},
  author={Bai, Shuai and Chen, Keqin and Liu, Xuejing and Wang, Jialin and Ge, Wenbin and Song, Sibo and Dang, Kai and Wang, Peng and Wang, Shijie and Tang, Jun and others},
  journal={arXiv preprint arXiv:2502.13923},
  year={2025}
}

@article{qwen3,
  title={Qwen3 technical report},
  author={Yang, An and Li, Anfeng and Yang, Baosong and Zhang, Beichen and Hui, Binyuan and Zheng, Bo and Yu, Bowen and Gao, Chang and Huang, Chengen and Lv, Chenxu and others},
  journal={arXiv preprint arXiv:2505.09388},
  year={2025}
}

@article{gpt4o,
  title={Gpt-4o system card},
  author={Hurst, Aaron and Lerer, Adam and Goucher, Adam P and Perelman, Adam and Ramesh, Aditya and Clark, Aidan and Ostrow, AJ and Welihinda, Akila and Hayes, Alan and Radford, Alec and others},
  journal={arXiv preprint arXiv:2410.21276},
  year={2024}
}

@inproceedings{docling,
  title={Docling: An Efficient Open-Source Toolkit for AI-driven Document Conversion},
  author={Livathinos, Nikos and Auer, Christoph and Lysak, Maxim and Nassar, Ahmed and Dolfi, Michele and Vagenas, Panos and Ramis, Cesar Berrospi and Omenetti, Matteo and Dinkla, Kasper and Kim, Yusik and others},
  booktitle={AAAI Conference on Artificial Intelligence},
  year={2025}
}

@article{gemini25,
  title={Gemini 2.5: Pushing the frontier with advanced reasoning, multimodality, long context, and next generation agentic capabilities},
  author={Comanici, Gheorghe and Bieber, Eric and Schaekermann, Mike and Pasupat, Ice and Sachdeva, Noveen and Dhillon, Inderjit and Blistein, Marcel and Ram, Ori and Zhang, Dan and Rosen, Evan and others},
  journal={arXiv preprint arXiv:2507.06261},
  year={2025}
}

@article{internvl35,
  title={Internvl3. 5: Advancing open-source multimodal models in versatility, reasoning, and efficiency},
  author={Wang, Weiyun and Gao, Zhangwei and Gu, Lixin and Pu, Hengjun and Cui, Long and Wei, Xingguang and Liu, Zhaoyang and Jing, Linglin and Ye, Shenglong and Shao, Jie and others},
  journal={arXiv preprint arXiv:2508.18265},
  year={2025}
}

@misc{nanonetsocr,
  title={Nanonets-OCR-S: A model for transforming documents into structured markdown with intelligent content recognition and semantic tagging},
  author={Souvik Mandal and Ashish Talewar and Paras Ahuja and Prathamesh Juvatkar},
  year={2025},
}

@misc{chandra,
  author       = {Datalab To},
  title        = {Chandra: OCR model that handles complex tables, forms, handwriting with full layout},
  year         = {2025},
  version      = {v0.1.7},
  url          = {https://github.com/datalab-to/chandra},
  note         = {Open-source code, Apache 2.0 license}
}

@inproceedings{teds-score,
  title={Image-based table recognition: data, model, and evaluation},
  author={Zhong, Xu and ShafieiBavani, Elaheh and Jimeno Yepes, Antonio},
  booktitle={European conference on computer vision},
  pages={564--580},
  year={2020},
  organization={Springer}
}

@inproceedings{li2025readoc,
  title={Readoc: A unified benchmark for realistic document structured extraction},
  author={Li, Zichao and Abulaiti, Aizier and Lu, Yaojie and Chen, Xuanang and Zheng, Jia and Lin, Hongyu and Han, Xianpei and Jiang, Shanshan and Dong, Bin and Sun, Le},
  booktitle={Findings of the Association for Computational Linguistics: ACL 2025},
  pages={21889--21905},
  year={2025}
}

@article{li2020docbank,
  title={Docbank: A benchmark dataset for document layout analysis},
  author={Li, Minghao and Xu, Yiheng and Cui, Lei and Huang, Shaohan and Wei, Furu and Li, Zhoujun and Zhou, Ming},
  journal={arXiv preprint arXiv:2006.01038},
  year={2020}
}

@article{li2022pp,
  title={Pp-structurev2: A stronger document analysis system},
  author={Li, Chenxia and Guo, Ruoyu and Zhou, Jun and An, Mengtao and Du, Yuning and Zhu, Lingfeng and Liu, Yi and Hu, Xiaoguang and Yu, Dianhai},
  journal={arXiv preprint arXiv:2210.05391},
  year={2022}
}

@inproceedings{donut,
  title={Ocr-free document understanding transformer},
  author={Kim, Geewook and Hong, Teakgyu and Yim, Moonbin and Nam, JeongYeon and Park, Jinyoung and Yim, Jinyeong and Hwang, Wonseok and Yun, Sangdoo and Han, Dongyoon and Park, Seunghyun},
  booktitle={European Conference on Computer Vision},
  pages={498--517},
  year={2022},
  organization={Springer}
}

@article{hu2022lora,
  title={Lora: Low-rank adaptation of large language models.},
  author={Hu, Edward J and Shen, Yelong and Wallis, Phillip and Allen-Zhu, Zeyuan and Li, Yuanzhi and Wang, Shean and Wang, Liang and Chen, Weizhu and others},
  journal={Iclr},
  volume={1},
  number={2},
  pages={3},
  year={2022}
}

@article{luo2025empirical,
  title={An empirical study of catastrophic forgetting in large language models during continual fine-tuning},
  author={Luo, Yun and Yang, Zhen and Meng, Fandong and Li, Yafu and Zhou, Jie and Zhang, Yue},
  journal={IEEE Transactions on Audio, Speech and Language Processing},
  year={2025},
  publisher={IEEE}
}

@article{chung2025finetuning,
  title={Finetuning Vision-Language Models as OCR Systems for Low-Resource Languages: A Case Study of Manchu},
  author={Chung, Yan Hon Michael and Choi, Donghyeok},
  journal={arXiv preprint arXiv:2507.06761},
  year={2025}
}

@inproceedings{kitabbench,
  title={Kitab-bench: A comprehensive multi-domain benchmark for arabic ocr and document understanding},
  author={Heakl, Ahmed and Sohail, Muhammad Abdullah and Ranjan, Mukul and Elbadry, Rania and Ahmad, Ghazi Shazan and El-Geish, Mohamed and Maher, Omar and Shen, Zhiqiang and Khan, Fahad Shahbaz and Khan, Salman},
  booktitle={Findings of the Association for Computational Linguistics: ACL 2025},
  pages={22006--22024},
  year={2025}
}

@article{dpo,
  title={Direct preference optimization: Your language model is secretly a reward model},
  author={Rafailov, Rafael and Sharma, Archit and Mitchell, Eric and Manning, Christopher D and Ermon, Stefano and Finn, Chelsea},
  journal={Advances in neural information processing systems},
  volume={36},
  pages={53728--53741},
  year={2023}
}

@article{zhu2025teach,
  title={How to Teach Large Multimodal Models New Skills},
  author={Zhu, Zhen and Gong, Yiming and Xiao, Yao and Liu, Yaoyao and Hoiem, Derek},
  journal={arXiv preprint arXiv:2510.08564},
  year={2025}
}

@article{opencv_library,
  author = {Bradski, G.},
  title = {{The OpenCV Library}},
  journal = {Dr. Dobb's Journal of Software Tools},
  year = {2000}
}

@misc{rocksdb2024,
  author = {{Meta Platforms, Inc.}},
  title = {{RocksDB}: A Persistent Key-Value Store for Fast Storage Environments},
  year = {2024},
  howpublished = {\url{https://github.com/facebook/rocksdb}},
  note = {Accessed: 2024}
}

@article{fasttext,
  title={Fasttext. zip: Compressing text classification models},
  author={Joulin, Armand and Grave, Edouard and Bojanowski, Piotr and Douze, Matthijs and J{\'e}gou, H{\'e}rve and Mikolov, Tomas},
  journal={arXiv preprint arXiv:1612.03651},
  year={2016}
}

@inproceedings{layoutlmv3,
  title={Layoutlmv3: Pre-training for document ai with unified text and image masking},
  author={Huang, Yupan and Lv, Tengchao and Cui, Lei and Lu, Yutong and Wei, Furu},
  booktitle={Proceedings of the 30th ACM international conference on multimedia},
  pages={4083--4091},
  year={2022}
}

@inproceedings{ccnet,
  title={CCNet: Extracting high quality monolingual datasets from web crawl data},
  author={Wenzek, Guillaume and Lachaux, Marie-Anne and Conneau, Alexis and Chaudhary, Vishrav and Guzm{\'a}n, Francisco and Joulin, Armand and Grave, Edouard},
  booktitle={Proceedings of the Twelfth Language Resources and Evaluation Conference},
  pages={4003--4012},
  year={2020}
}

@inproceedings{resnet,
  title={Deep residual learning for image recognition},
  author={He, Kaiming and Zhang, Xiangyu and Ren, Shaoqing and Sun, Jian},
  booktitle={Proceedings of the IEEE conference on computer vision and pattern recognition},
  pages={770--778},
  year={2016}
}

@article{faiss,
  title={The faiss library},
  author={Douze, Matthijs and Guzhva, Alexandr and Deng, Chengqi and Johnson, Jeff and Szilvasy, Gergely and Mazar{\'e}, Pierre-Emmanuel and Lomeli, Maria and Hosseini, Lucas and J{\'e}gou, Herv{\'e}},
  journal={IEEE Transactions on Big Data},
  year={2025},
  publisher={IEEE}
}

@article{nagel2023common,
  title={Common Crawl: Data collection and use cases for NLP},
  author={Nagel, Sebastian},
  journal={HPLT \& NLPL Winter School on Large-Scale Language Modeling and Neural Machine Translation with Web Data, February},
  volume={6},
  year={2023}
}

@article{weber2023wordscape,
  title={WordScape: a Pipeline to extract multilingual, visually rich Documents with Layout Annotations from Web Crawl Data},
  author={Weber, Maurice and Siebenschuh, Carlo and Butler, Rory and Alexandrov, Anton and Thanner, Valdemar and Tsolakis, Georgios and Jabbar, Haris and Foster, Ian and Li, Bo and Stevens, Rick and others},
  journal={Advances in Neural Information Processing Systems},
  volume={36},
  pages={26048--26068},
  year={2023}
}

@inproceedings{dit700,
  title={From chaotic ocr words to coherent document: A fine-to-coarse zoom-out network for complex-layout document image translation},
  author={Zhang, Zhiyang and Zhang, Yaping and Liang, Yupu and Xiang, Lu and Zhao, Yang and Zhou, Yu and Zong, Chengqing},
  booktitle={Proceedings of the 31st International Conference on Computational Linguistics},
  pages={10877--10890},
  year={2025}
}
